\documentclass[pmlr,twocolumn,10pt]{jmlr} 

\usepackage{algpseudocode}

\usepackage{stackengine}
\usepackage{booktabs}
\usepackage{multirow, multicol}
\usepackage{siunitx}

\newcommand{\equal}[1]{{\hypersetup{linkcolor=black}\thanks{#1}}}

\newcommand{\mname}{\texttt{FedPxN}\xspace}
\theorembodyfont{\upshape}
\theoremheaderfont{\scshape}
\theorempostheader{:}
\theoremsep{\newline}

\jmlrvolume{LEAVE UNSET}
\jmlryear{2023}
\jmlrsubmitted{LEAVE UNSET}
\jmlrpublished{LEAVE UNSET}
\jmlrworkshop{Conference on Health, Inference, and Learning (CHIL) 2023} 

\title[Towards the Practical Utility of FL in the Medical Domain]{
Towards the Practical Utility of \\Federated Learning in the Medical Domain}

\author{%
\Name{Seongjun Yang}\equal{These authors contributed equally}\equal{Work done at KAIST} \Email{seongjunyang@krafton.com}\\
\addr KRAFTON, Republic of Korea
\AND
\Name{Hyeonji Hwang}\footnotemark[1] \Email{localh@kaist.ac.kr}\\
\Name{Daeyoung Kim} \Email{daeyoung.k@kaist.ac.kr}\\
\addr KAIST, Republic of Korea 
\AND
\Name{Radhika Dua}\footnotemark[2] \Email{radhikadua@google.com}\\
\addr Google Research, India
\AND
\Name{Jong-Yeup Kim} \Email{jykim@kyuh.ac.kr}\\
\addr College of Medicine, Konyang University, Republic of Korea
\AND
\Name{Eunho Yang} \Email{eunhoy@kaist.ac.kr}\\
\Name{Edward Choi} \Email{edwardchoi@kaist.ac.kr}\\
\addr KAIST, Republic of Korea
}

\begin{document}

\maketitle

\begin{abstract}
Federated learning (FL) is an active area of research. One of the most suitable areas for adopting FL is the medical domain, where patient privacy must be respected. Previous research, however, does not provide a practical guide to applying FL in the medical domain. We propose empirical benchmarks and experimental settings for three representative medical datasets with different modalities: longitudinal electronic health records, skin cancer images, and electrocardiogram signals. The likely users of FL such as medical institutions and IT companies can take these benchmarks as guides for adopting FL and minimize their trial and error. For each dataset, each client data is from a different source to preserve real-world heterogeneity. We evaluate six FL algorithms designed for addressing data heterogeneity among clients, and a hybrid algorithm combining the strengths of two representative FL algorithms. Based on experiment results from three modalities, we discover that simple FL algorithms tend to outperform more sophisticated ones, while the hybrid algorithm consistently shows good, if not the best performance. We also find that a frequent global model update leads to better performance under a fixed training iteration budget. As the number of participating clients increases, higher cost is incurred due to increased IT administrators and GPUs, but the performance consistently increases. We expect future users will refer to these empirical benchmarks to design the FL experiments in the medical domain considering their clinical tasks and obtain stronger performance with lower costs.
\end{abstract}

\paragraph*{Data and Code Availability}
Every detail of the data availability and code for every experiment in this study is stated on the official repository. 
\footnote{\url{https://github.com/wns823/medical_federated}}
For eICU dataset, completing the CITI “Data or Specimens Only Research” course and a formal request are necessary.

\section{Introduction}
\label{sec:intro}
Federated learning (FL) is a distributed machine learning framework in which each client does not share its data but instead shares model parameters, thus preserving data privacy. FL is divided into cross-device and cross-silo FL \citep{kairouz2021advances, wang2021field}. Typically, mobile devices are the clients in a cross-device setting, and hence the Internet connectivity and the efficiency of training in each device are the critical factors. On the other hand, the well-known example of the cross-silo FL is in the medical domain where medical institutions are participating. There are relatively fewer clients and the Internet connectivity is not critical due to LAN connections \citep{kairouz2021advances}. Recently, researchers have begun to test FL on medical datasets with varying success \citep{pfohl2019federated, brisimi2018federated, dou2021federated, boughorbel2019federated, liu2018fadl, sheller2020federated}. However, existing studies do not fully consider the real-world situation of adopting FL in the medical domain. 

The main stakeholders will be hospitals and IT service providers. Hospitals want to take advantage of machine learning models using their large data for various reasons; to use the models for clinical decision-making or to increase efficiency. FL enables multiple distributed data holders to collaboratively train a shared model and get generalizable AI models \citep{10.1093/jamia/ocac188}. The hospital can take advantage of the knowledge from other hospitals, especially for unseen data. Large IT companies or startups are not able to reach the medical dataset due to privacy issues. They can develop practical machine learning models with real patient records via FL. Achieving high performance at a low monetary cost is an important concern of these likely users. 

However, applying FL involves a lot of considerations other than a model architecture or data construction due to its complicated process (i.e. local training and model aggregation). Moreover, it is more expensive than single-site learning to change the settings as there are many clients related and communication rounds are needed to get a satisfactory result. The first consideration is that they have to choose an appropriate FL algorithm. There are various modalities of data from the medical institutions: structured, image, or signal data. A suitable algorithm for each modality may vary and this area has not been fully explored. In addition to choosing which algorithm to use, there are still many other decisions waiting for them to make; the number of training epochs in each client server, the type of normalization technique to use, and the number of participating clients. They should also consider the approximate power consumption of the FL framework because it is directly related to the monetary cost. The guidance for these FL settings will decrease the expense of trial and error and help the users to concentrate on better model architecture or data strategy.

In this work, for the first time, we test well-known FL algorithms on three representative real-world medical datasets with different modalities involving structured (i.e., tabular), visual, and signal data. To be more realistic, each dataset is from a different source respectively so the data distributions preserve real-world heterogeneity. We select FL algorithms designed for heterogeneous data distributions among clients to observe whether they solve data heterogeneity generated by real-world healthcare applications. We provide practical benchmarks including the normalization techniques and the number of local training epochs. We also evaluate the performance of each FL algorithm in terms of monetary costs, such as power consumption and the number of participating clients.
We combine two FL algorithms and test the hybrid method (\mname) on all three settings. \mname constantly shows comparable or better performance compared to other methods. We recommend using this version to minimize the expense of trial and error in choosing an FL algorithm.












\section{Related Works}
\paragraph{FL algorithms on data that are not independent and identically distributed (i.i.d.)} The main challenge of FL in the medical domain is a non-i.i.d. problem \citep{rieke2020future, li2022federated} because of factors such as different specific protocols, medical devices, and local demographics. FedAvg \citep{mcmahan2017communication} is a widely known framework in FL but does not ensure training convergence when data are heterogeneous over local clients \citep{li2019convergence, hsu2019measuring}. 
Therefore, a large number of methods focus on addressing the non-i.i.d. problem. FedProx \citep{li2020federated} adds a proximal term in the local objective of the FedAvg framework to solve the heterogeneity problem, and FedOpt \citep{reddi2020adaptive} applies adaptive optimization in global aggregation to stabilize convergence for heterogeneous data. The inconsistency of local and global objectives due to the data heterogeneity of each local client is handled in Scaffold \citep{karimireddy2020scaffold} and FedDyn \citep{acar2021federated}. Scaffold computes and aggregates control variates, whereas FedDyn uses a dynamic regularizer to solve the problem. Most FL methods \citep{mcmahan2017communication, li2020federated, reddi2020adaptive, karimireddy2020scaffold, acar2021federated} are validated in a label-heterogeneous experimental setting by partitioning the same sourced dataset into multiple clients. In contrast, FedBN \citep{li2021fedbn} considered the non-i.i.d problem that can occur due to feature shifts in different data sources, and proposed to aggregate local models without batch normalization layers to handle this problem.
Similarly, SiloBN \citep{andreux2020siloed} aggregates local models without local batch normalization statistics. FedDAR \citep{zhong2023feddar} decouples domain-specific prediction heads and a shared encoder in order to tackle a non-i.i.d setting where there is a similarity between each domain.
\paragraph{FL in Healthcare} The medical domain is an active area of FL nowadays because of the importance of patient privacy. Prior study on FL in the medical domain \citep{lee2020federated} has achieved performance comparable to that of centralized learning. However, these studies focused only on electronic health records (EHR), and electrocardiogram (ECG) signals. Moreover, they randomly sample the data to create heterogeneity. However, in reality, heterogeneity also exists in the feature space. In our work, we focus on diverse and important modalities in healthcare (images, EHR, and ECG) and also consider a realistic scenario in which the clients are from different hospitals. Effective FL frameworks for specific tasks have been empirically demonstrated for each specific modality such as EHR and medical images in 
\citep{huang2019patient, kim2017federated, liu2019two, park2021federated, xu2020federated}. However, these results depended on architecture or were validated only on specific modalities.
\paragraph{Normalization layer for FL} In \citep{diao2020heterofl, hsieh2020non}, the authors compared performance when using various normalization layers in the local model for image classification tasks. 
Our study observes which normalization is effective in eight clinical tasks.

\section{Methods}
FL methods are often designed to minimize the weighted average of local objective function of clients as follows:
\begin{equation}
\label{eq:loss}
\min_{w}F(w):=\sum_{k=1}^{K}\frac{n_k}{n}F_{k}(w),
\end{equation}
where $K$ is the number of clients, $n_k$ the number of examples in each client $k$, $n$ the total data size of all clients, and $F_k$ the local objective function of each client $k$. Data heterogeneity is a key and common challenge in solving Eq.~\ref{eq:loss}. In this section, we propose a hybrid FL method (\mname) based on the analysis of the different training patterns of the parameters of each client in FedBN \citep{li2021fedbn} and deployment of the proximal term \citep{li2020federated}.

\subsection{FedAvg}
FedAvg \citep{mcmahan2017communication} is the de facto standard algorithm in FL (Appendix Algorithm \ref{algo:fedavg}). In this framework, first, the central server sends the global model $w_{t}$ to the clients in each communication round $t$. Then, each client $k$ sends the updated model $w_{t,k}$ back to the server after local training. Next, the central server averages all client models considering the data size ratio $n_k / n$. Also, FedAvg reduces the number of communication rounds required for model convergence by updating the model after multiple epochs $E$ of local training. Every baseline of our work is based on FedAvg, so we use the notation used in FedAvg for describing other methods.

\subsection{FedProx}
To solve the non-i.i.d problem, FedProx \citep{li2020federated} introduces an $L_2$ regularization term $\|w_{t,k} - w_{t}\|^2_2$ to the local objective function $F_{k}$ of the FedAvg framework as follows:
\begin{equation}
\small
\label{eq:fedprox}
\tilde{F}_k(w_{t,k}; b) = F_k(w_{t,k};b) + \frac{\mu}{2} \| w_{t,k} - w_{t}  \|^2_2,
\end{equation}
where $F$ is the objective function and $\mu$ is a hyperparameter that controls the degree of regularization.
The local model updates are restricted by the regularization term so that they are closer to the global model.

\vspace{-1mm}
\subsection{FedBN}
\vspace{-1mm}
In FedBN \citep{li2021fedbn}, the local models are aggregated without batch normalization layers in order to handle feature shifts among clients. The study evaluated FedBN's effectiveness through experiments using datasets from different sources.
In our study, we use different types of normalization (i.e., batch normalization (BN) \citep{ioffe2015batch}, group normalization (GN) \citep{wu2018group}, and layer normalization (LN) \citep{ba2016layer} for each task to maximize the performance of the models.
Hence, we extend FedBN in such a way that all layers of the local models except the normalization layers are aggregated.

\subsection{Hybrid Algorithm: \mname}
\vspace{-2mm}
Because the statistics used in the normalization layers are different in each local client $w_{t,k}$, it could be challenging for the aggregated model $w_{t}$ to capture the distribution of data collected from different sources.
A simple solution is to aggregate the local client models without the normalization layers, as in FedBN.

\begin{figure}[h!]
\floatconts
{fig:normalization}
{\caption{$L_2$ distance between each client's model parameters and the global parameters, excluding the normalization layers, when using (a) FedBN and (b) \mname. Both show results of the mortality prediction task, in which all models were equipped with LN, and trained using the five clients with the largest samples. }}
{
    \centering
    \subfigure[FedBN]{\label{fig:norm-a}
      \includegraphics[width=0.5\columnwidth]{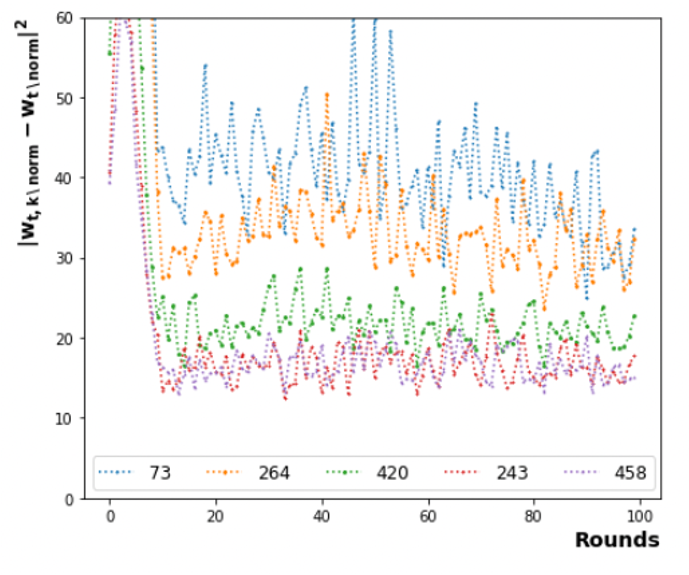}}
    \hspace{-4.5mm}
    \subfigure[\mname]{\label{fig:norm-b}
      \includegraphics[width=0.5\columnwidth]{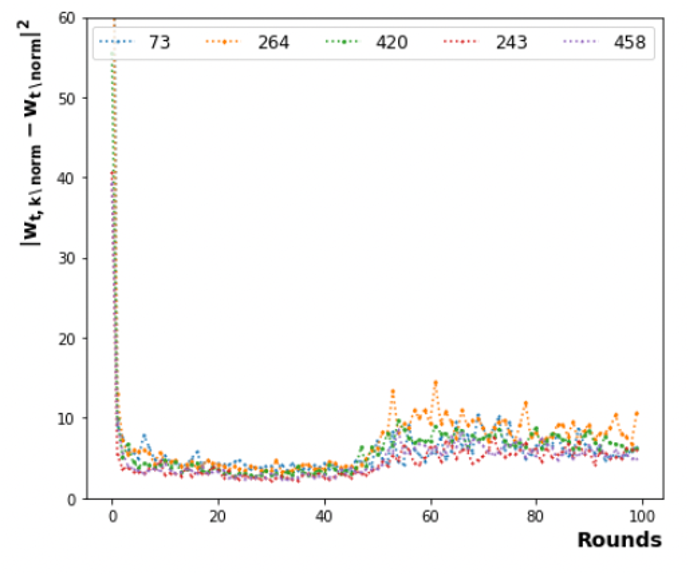}}
}
\vspace{-3mm}
\end{figure}

The normalization layers in FedBN naturally move towards their own local optimum depending on the data distribution of each client.
However, we suspect that the other non-normalization layers might also move towards the local optimum inconsistent with the global optimum because of the effect of the normalization layers in the model of each client.
Therefore, we measure the $L_2$ distance between all $w_{t,k \setminus norm}$,
and $w_{t \setminus norm}$,
where the former are the local model parameters except for the normalization layers in client $k$,
and the latter are the global model parameters except the normalization layers in each round $t$, as follows:
\vspace{-1mm}
\begin{equation}
\vspace{-1mm}
\small
\label{eq:l2distance}
\|w_{t,k \setminus norm} - w_{t \setminus norm}\|^2_2,
\vspace{-1mm}
\end{equation}
where $t$ indicates each communication round, $k$ denotes each local client.
As shown on \figureref{fig:norm-a}, each client's model $w_{t,k \setminus norm}$ has evolved in different directions from the global model $w_{t \setminus norm}$. We hypothesize that a proximal term (Equation \ref{eq:l2distance}) inspired by FedProx will help the local models in FedBN to progress towards the global optimum. 
As \figureref{fig:norm-b} shows, the added proximal term prevents $w_{t,k \setminus norm}$ from deviating too much from $w_{t \setminus norm}$.

Based on this observation, we propose \mname, an algorithm encouraging the normalization layers to adapt to each client's unique feature distribution and the other layers to follow the global optimum in order to improve the overall performance.
In \mname, local models are updated by the local objective with the proximal term of the other layers of the global model during local training and then aggregated without the normalization layers.
The detailed algorithm is presented in the Appendix Algorithm~\ref{algo:fedpxn}.

\vspace{-4mm}
\section{Experimental Setup}
In this section, we introduce the experimental settings and different medical datasets used to evaluate all FL frameworks for the medical domain in terms of their practical utility.
First of all, we compare our approach with recent well-known FL methods for solving the non-i.i.d problem including FedAvg \citep{mcmahan2017communication}, FedProx \citep{li2020federated}, FedBN \citep{li2021fedbn}, FedOpt (FedAdam, FedAdagrad, FedYoGi) \citep{reddi2020adaptive}, and FedDyn \citep{acar2021federated}. More information about FL methods is described in Appendix \ref{appendix:explain_methods}.

For all our experiments, we assumed the full participation of clients in all communication rounds by LAN connection.
For all FL methods in each of the tasks in our study, we used the same deep learning model, data size, input size, and GPU in order to conduct a fair comparison. The CPU spec for our experimental setting is AMD EPYC 7502 32-Core Processor(2.5GHz). We used NVIDIA GeForce RTX 3090 with a RAM size of 24G and the corresponding CUDA version was 11.4. Further, we fixed the total number of training epochs for each local client in all FL methods, where this number is defined as the number of local epochs times the number of communication rounds. 
If not specified, our default setting for local epochs is a single epoch. We also conducted the experiments by setting various communication rounds and local epochs. More details are described in Appendix~\ref{ex:details}. Also, Experiment results about various combinations of communication rounds and local epochs are discussed in Section~\ref{ssec:local_epoch}. 

\begin{figure*}[h!]
\vspace{-3mm}
\floatconts
{fig:labeldist}

\centering
{
    \subfigure[eICU Acuity]{\label{fig:label-eicu}
      \includegraphics[width=0.65\columnwidth]{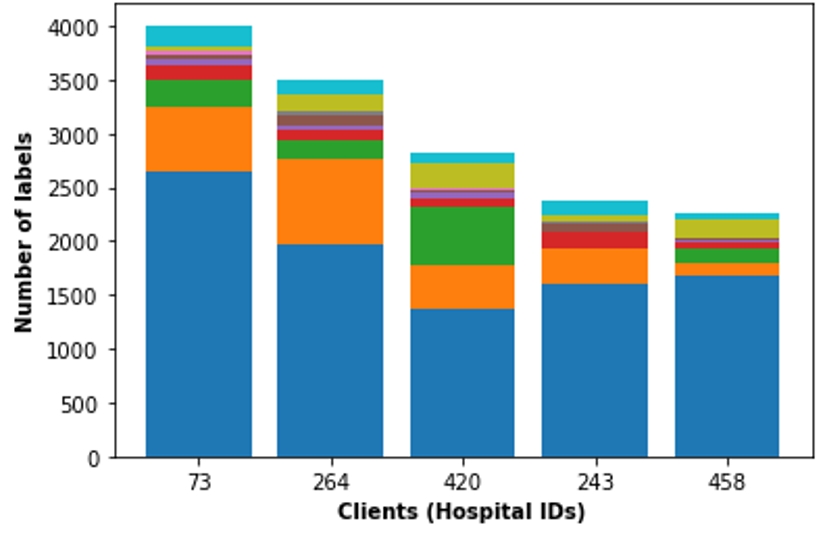}}
    \subfigure[skin cancer datasets]{\label{fig:label-skin}
      \includegraphics[width=0.65\columnwidth]{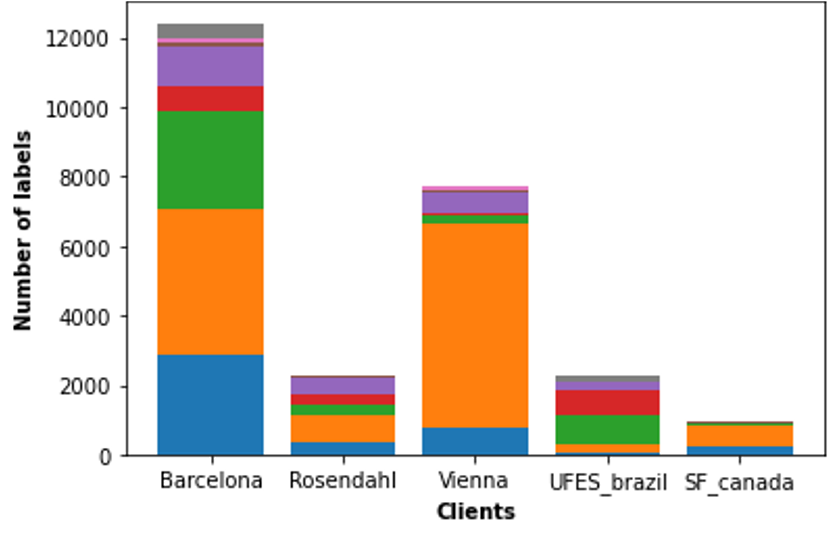}}
    \subfigure[ECG datasets]{\label{fig:label-ecg}
      \includegraphics[width=0.65\columnwidth]{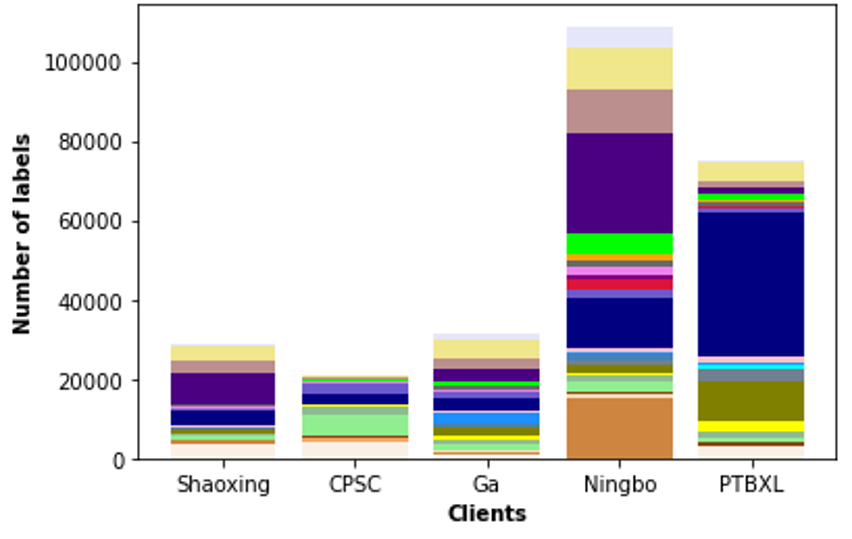}}
}
{\caption{Label distributions of eICU, skin cancer, and ECG datasets. Colors indicate the labels for each client. The label distribution shown for eICU is for the final acuity prediction task for the five clients with the largest datasets. \vspace{-2.5mm}}}
\end{figure*}

\vspace{-3mm}
\subsection{Electronic Health Records}
\label{sec:exp_setup_ehr}
\vspace{-2mm}
We first evaluated FL methods on the eICU dataset~\citep{pollard2018eicu} by taking advantage of an available benchmark dataset~\citep{mcdermott2021comprehensive}, which consists of intensive care unit (ICU) records of patients aged $15$ years or over. The benchmark dataset contains 71477 ICU stays
across 59 hospitals, where ICU stays range between 540 and 4008 for each hospital. To conduct experiments in the FL setting, we used the $5$, $10$, $20$, and $30$ hospitals with the most ICU stays from the dataset and performed the six clinical prediction tasks; two mortality prediction tasks (mort\_24h, mort\_48h), two discharge perdiction tasks (disch\_24h, disch\_48h), length of stay prediction (LOS), and final acuity prediction (Acuity). Mort\_24h, mort\_48h, and LOS are binary classification tasks. Disch\_24h, disch\_48h, and Acuity are 10-way classification tasks. More details about the clinical tasks are in Appendix \ref{appendix:eicu_info}.

Among all tasks that have heterogeneous label distribution, as an example, we show the label distribution of the Final acuity prediction task in \figureref{fig:label-eicu}.
While the most dominant label is shared by all clients, the label distributions and data volumes vary across the clients.
We used a Transformer classifier \citep{vaswani2017attention} with LN or GN in all tasks. Full details are given in Appendix \ref{appendix:eicu_info}. 
We utilized the AUROC and AUPRC scores to evaluate the performance of the trained model.

\vspace{-2.5mm}
\subsection{Skin cancer image dataset}
We also evaluated all FL methods on skin cancer datasets. To be more specific, we constructed five clients which are from different sources (Barcelona, Vienna, Rosendahl, UFES$\_$brazil, SF$\_$canada). The details of data construction are illustrated in Appendix~\ref{appendix:skin_info}. We observe that the dominant label is almost the same for all clients and all clients have an imbalanced dataset. Further, we observe that the second dominant label is more different for each client, and the total number of samples significantly varies from one client to another. The details of the label distribution and size of the data for each client are shown in \figureref{fig:label-skin}. 

Then, we formulated an 8-way classification task that uses an image as input and predicts the type of skin cancer. Also, we used an EfficientNet-B0 \citep{tan2019efficientnet} with BN or GN. To evaluate the performance of the model, we use two metrics, namely AUROC and AUPRC.  More details are also listed in Appendix \ref{appendix:skin_info}.

\vspace{-3mm}
\subsection{Electrocardiogram (ECG) dataset}
\vspace{-1mm}
We also evaluated the FL methods on ECG signals. We used the PhysioNet 2021 \citep{reyna2021will} dataset and split it into five clients based on the different hospitals (Shaoxing, CPSC, Ga, Ningbo, PTBXL). There are $26$ labels related to cardiac diseases. In \figureref{fig:label-ecg}, the label distribution of different clients for these data reveals that the total number of samples varies substantially from one client to another, and the data are imbalanced. Moreover, the dominant label in the data from each client varies. 

Our objective is to solve a multi-label prediction task to diagnose $26$ types of cardiac diseases in which a 12-lead ECG sample is given as input. We trained a ResNet-NC-SE \citep{kang2022study} with BN or GN because it showed the best performance on 12-lead ECG readings in PhysioNet 2021 to the best of our knowledge \citep{torch_ecg_paper, torch_ecg}. Following \cite{oh2022lead}, we evaluated the performance of our approach by measuring the CinC score, the official evaluation metric used by the Physionet 2021 challenge that takes into account domain knowledge on cardiovascular diseases. For more details, please refer to Appendix \ref{appendix:ecg_info}.

\begin{table*}[h!]
\caption{AUROC results for the eICU dataset using the data of the five largest clients. For each FL method, bold indicates the better normalization technique (LN or GN). We indicate the highest average AUROC results for all six tasks in blue. }
\label{tab:eicu_5_clients_average}
\resizebox{\textwidth}{!}{
\begin{tabular}{c|ccccccccc} 
\toprule
\multicolumn{1}{l}{}              & \multicolumn{1}{l}{} & FedAvg                  & FedProx                 & FedBN                 & FedAdam                 & FedAdagrad              & FedYoGi                 & FedDyn                  & \mname                          \\ 
\hline
\multirow{2}{*}{mort\_24h}        & LN                   & 67.85$\pm$2.52          & 73.29$\pm$2.13          & \textbf{69.31$\pm$0.56} & \textbf{68.27$\pm$1.98} & \textbf{67.76$\pm$1.85} & 60.75$\pm$3.78          & \textbf{70.80$\pm$0.68} & 74.05$\pm$1.61                    \\
                                  & GN                   & \textbf{69.54$\pm$2.23} & \textbf{74.02$\pm$2.57} & 69.05$\pm$1.64          & 67.60$\pm$2.39          & 67.21$\pm$0.77          & \textbf{70.82$\pm$3.57} & 69.79$\pm$2.09          & \textbf{75.07$\pm$1.83}           \\ 
\hline
\multirow{2}{*}{mort\_48h}        & LN                   & \textbf{68.33$\pm$3.55} & 72.37$\pm$1.42          & 68.38$\pm$0.81          & 69.61$\pm$1.31          & \textbf{68.62$\pm$1.47} & 68.69$\pm$0.56          & 67.99$\pm$1.23          & 72.22$\pm$1.87                    \\
                                  & GN                   & 67.81$\pm$0.58          & \textbf{72.46$\pm$1.42} & \textbf{69.42$\pm$1.08} & \textbf{70.71$\pm$0.64} & 68.32$\pm$0.51          & \textbf{70.23$\pm$0.86} & \textbf{72.64$\pm$3.11} & \textbf{72.33$\pm$1.02}           \\ 
\hline
\multirow{2}{*}{LOS}              & LN                   & 63.23$\pm$0.31          & 62.97$\pm$0.34          & 63.04$\pm$0.13          & \textbf{62.46$\pm$0.40} & 62.61$\pm$0.57          & 62.31$\pm$0.71          & 61.97$\pm$0.53          & \textbf{63.73$\pm$0.42}           \\
                                  & GN                   & \textbf{63.61$\pm$0.24} & \textbf{63.55$\pm$0.27} & \textbf{63.36$\pm$0.03} & 62.36$\pm$0.85          & \textbf{62.83$\pm$0.65} & \textbf{62.54$\pm$0.98} & \textbf{63.05$\pm$1.11} & 63.68$\pm$0.18                    \\ 
\hline
\multirow{2}{*}{disch\_24h}       & LN                   & \textbf{67.58$\pm$0.40} & 67.95$\pm$0.56          & \textbf{68.18$\pm$0.22} & \textbf{66.37$\pm$1.42} & 66.50$\pm$0.32          & 66.26$\pm$1.02          & 66.99$\pm$0.87          & 67.98$\pm$0.57                    \\ 
                                  & GN                   & 67.38$\pm$0.61          & \textbf{68.13$\pm$0.34} & 67.63$\pm$0.56          & 66.34$\pm$1.39          & \textbf{68.31$\pm$0.76} & \textbf{66.89$\pm$0.78} & \textbf{68.01$\pm$0.95} & \textbf{68.49$\pm$0.39}           \\ 
\hline
\multirow{2}{*}{disch\_48h}       & LN                   & \textbf{68.72$\pm$0.80} & 68.49$\pm$0.97          & 69.37$\pm$0.58          & \textbf{68.59$\pm$0.70} & 68.20$\pm$0.59          & \textbf{68.39$\pm$0.81} & \textbf{69.51$\pm$0.73} & 68.51$\pm$0.22                    \\
                                  & GN                   & 67.90$\pm$1.19          & \textbf{69.01$\pm$1.00} & \textbf{69.43$\pm$1.32} & 68.23$\pm$0.55          & \textbf{68.23$\pm$0.27} & 68.09$\pm$0.54          & 69.31$\pm$1.52          & \textbf{68.79$\pm$0.59}           \\ 
\hline
\multirow{2}{*}{Acuity}           & LN                   & 71.60$\pm$0.20          & 71.56$\pm$0.09          & 71.99$\pm$0.34          & \textbf{70.70$\pm$0.78} & \textbf{70.58$\pm$0.79} & \textbf{69.40$\pm$0.87} & 70.74$\pm$0.48          & 71.40$\pm$0.35                    \\
                                  & GN                   & \textbf{71.79$\pm$0.04} & \textbf{71.94$\pm$0.30} & \textbf{72.16$\pm$0.23} & 69.24$\pm$0.69          & 68.97$\pm$0.37          & 69.15$\pm$0.44          & \textbf{71.43$\pm$0.58} & \textbf{72.01$\pm$0.83}           \\ 
\hline
\multirow{2}{*}{\textbf{Average}} & LN                   & 67.88                   & 69.44                   & 68.38                   & \textbf{67.67}          & \textbf{67.38}          & 65.97                   & 68                      & 69.65                             \\
                                  & GN                   & \textbf{68.01}          & \textbf{69.85}          & \textbf{68.51}          & 67.42                   & 67.31                   & \textbf{67.96}          & \textbf{69.04}          & \textbf{\textcolor{blue}{70.06}}  \\
\bottomrule
\end{tabular}
}
\end{table*}

\begin{figure*}[h!]
\centering
    {
        \subfigure[mort24h]{
          \includegraphics[width=0.68\columnwidth]{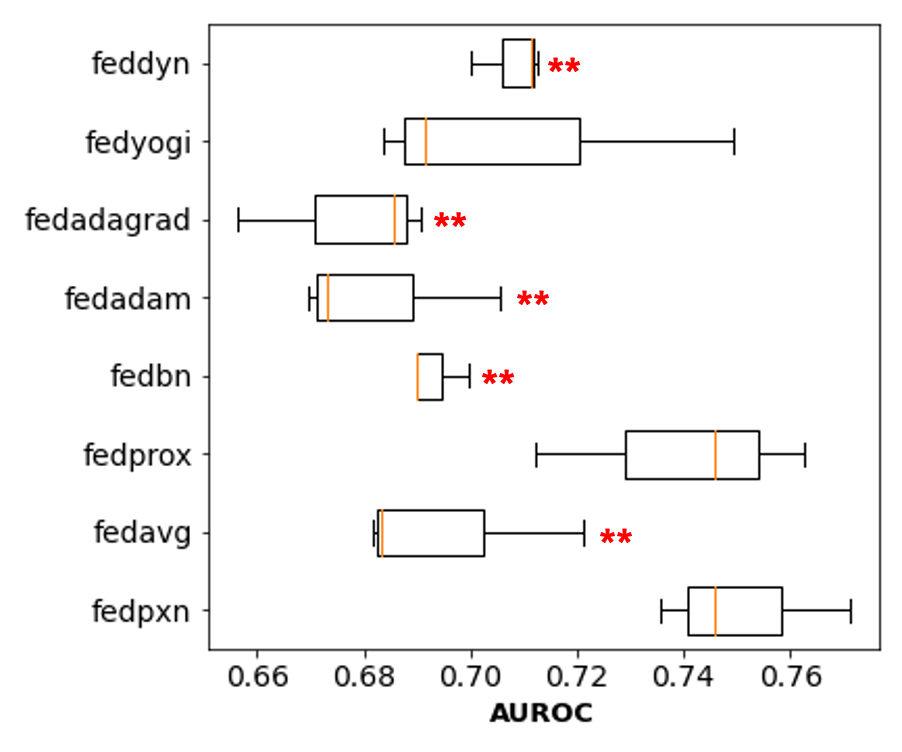}}
          \hspace{-4mm}
        \subfigure[mort48h]{
          \includegraphics[width=0.68\columnwidth]{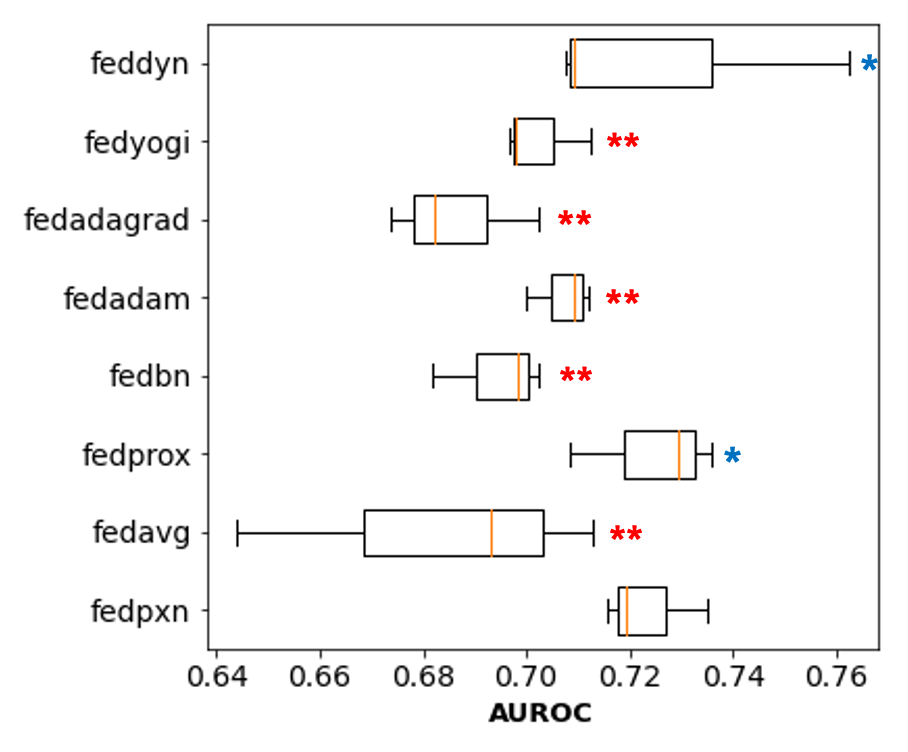}}
          \hspace{-4mm}
        \subfigure[LOS]{
          \includegraphics[width=0.68\columnwidth]{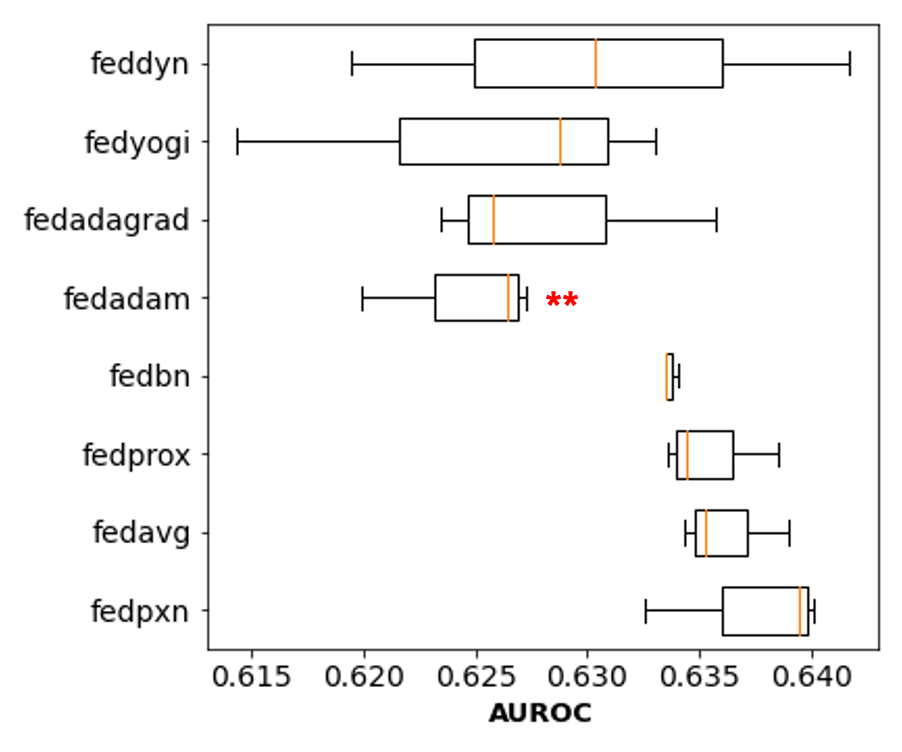}}
        \subfigure[disch24h]{
          \includegraphics[width=0.68\columnwidth]{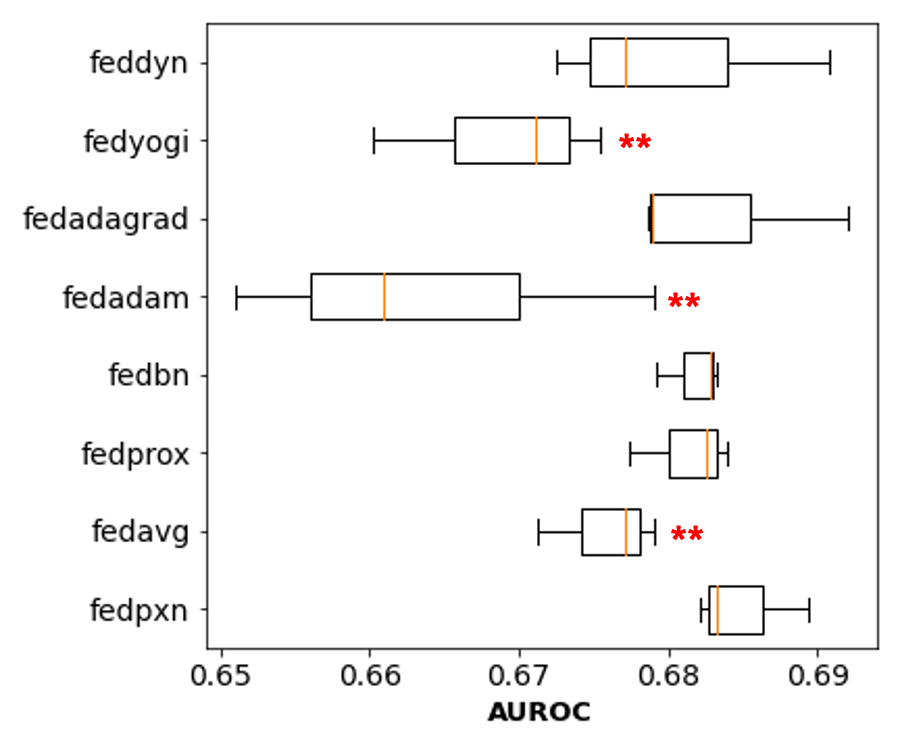}}
          \hspace{-4mm}
        \subfigure[disch48h]{
          \includegraphics[width=0.68\columnwidth]{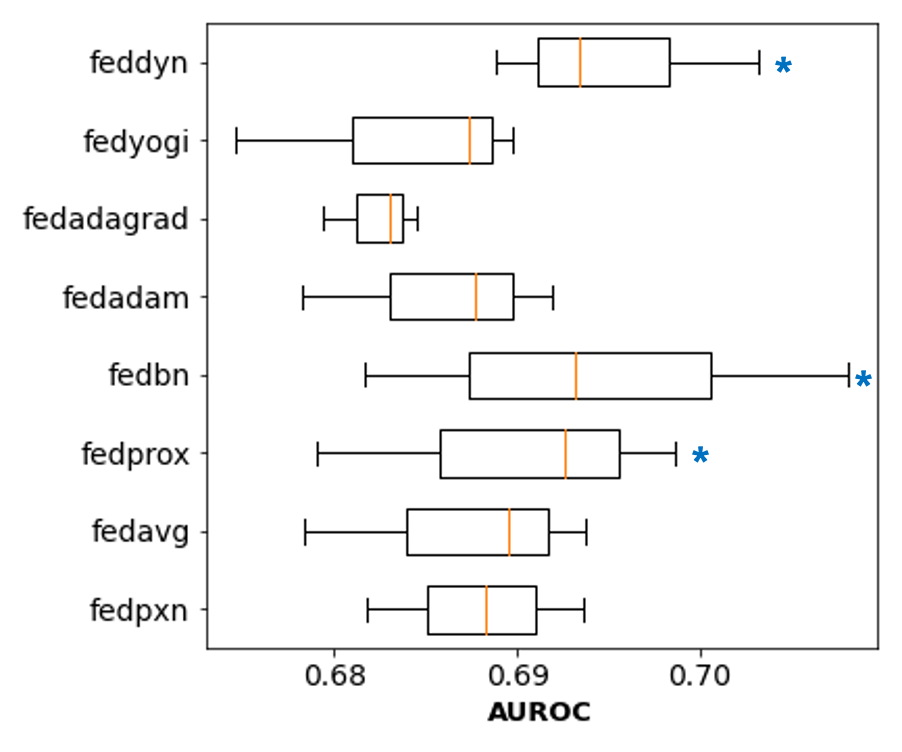}}
          \hspace{-4mm}
        \subfigure[Acuity]{
          \includegraphics[width=0.68\columnwidth]{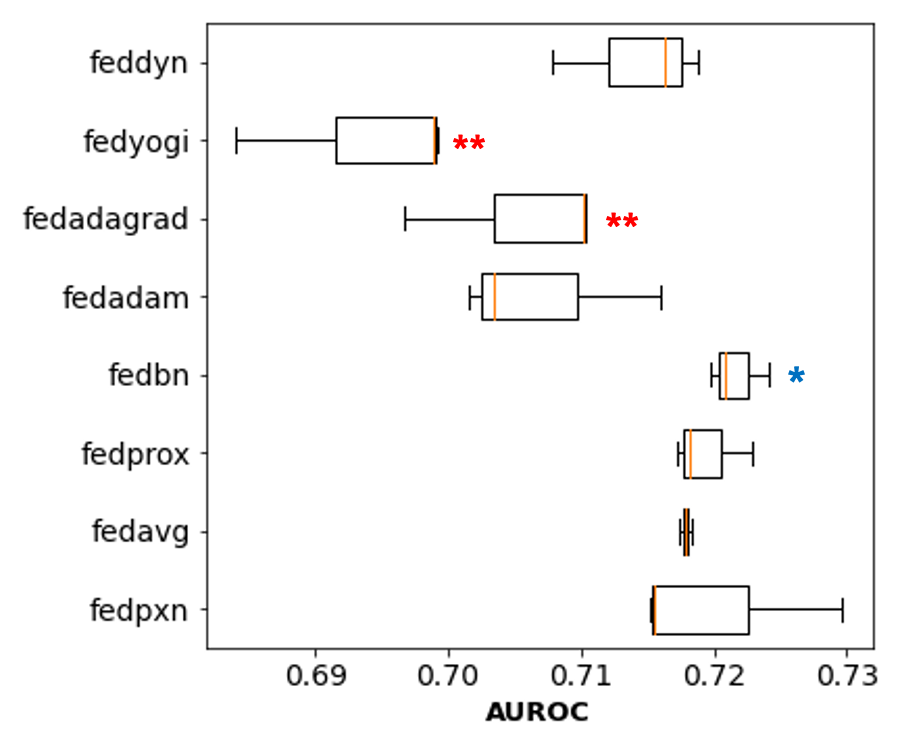}}      
    
\caption{Results of the statistical test with eICU datsets in the 5 clients setting. Two red astericks(**) show that FedPxN wins significantly(i.e. p-value is less than 0.05). One blue asterick(*) shows that FedPxN loses with average score, but insignificant(i.e. p-value is over 0.05). \mname never loses significantly. \vspace{-2.5mm}}
\label{fig:eicu_stat}}
\end{figure*}

\begin{table*}
\centering
\caption{AUROC and AUPRC results for the skin cancer dataset. Bold indicates the best normalization strategy for each method. We indicate the best performance in blue.}
\label{tab:skin_cancer_average}
\vspace{-2mm}
\resizebox{\textwidth}{!}{
\begin{tabular}{c|cccccc|c|ccccc|c} 
\toprule
\multicolumn{1}{c}{}        & \multicolumn{1}{l}{} & \multicolumn{6}{c|}{AUROC}                                                                                                     & \multicolumn{6}{c}{AUPRC}                                                                                                       \\ 
\hline
\multicolumn{1}{c}{}        &                      & Barcelona      & Rosendahl      & Vienna         & UFES\_brazil   & SF\_canada     & Avg                                       & Barcelona      & Rosendahl      & Vienna         & UFES\_brazil   & SF\_canada     & Avg                                        \\ 
\hline
\multirow{2}{*}{FedAvg}     & BN                   & 95.19$\pm$0.75 & 88.44$\pm$0.50 & 96.84$\pm$0.36 & 76.40$\pm$1.79 & 84.20$\pm$1.00 & \textbf{88.19$\pm$0.03}                   & 53.40$\pm$5.33 & 36.25$\pm$2.07 & 56.51$\pm$5.15 & 25.81$\pm$0.95 & 22.93$\pm$2.51 & \textbf{39.12$\pm$1.24}                    \\
                            & GN                   & 90.78$\pm$0.51 & 84.50$\pm$1.37 & 94.63$\pm$0.74 & 72.72$\pm$1.33 & 76.24$\pm$1.46 & 83.78$\pm$0.29                            & 35.81$\pm$3.92 & 29.91$\pm$1.60 & 39.59$\pm$2.50 & 23.36$\pm$0.97 & 20.06$\pm$1.07 & 29.75$\pm$1.66                             \\ 
\hline
\multirow{2}{*}{FedProx}    & BN                   & 95.82$\pm$0.43 & 87.56$\pm$1.61 & 96.40$\pm$0.93 & 76.79$\pm$0.06 & 85.26$\pm$3.19 & \textbf{88.37$\pm$0.68}                   & 58.63$\pm$2.64 & 36.47$\pm$3.38 & 61.80$\pm$4.76 & 25.92$\pm$0.55 & 23.36$\pm$1.28 & \textbf{41.24$\pm$1.33}                    \\
                            & GN                   & 90.28$\pm$0.24 & 82.70$\pm$1.26 & 94.51$\pm$0.45 & 71.88$\pm$1.75 & 73.58$\pm$2.71 & 82.59$\pm$0.25                            & 32.54$\pm$2.56 & 30.01$\pm$2.07 & 40.11$\pm$1.52 & 23.86$\pm$0.44 & 19.79$\pm$1.09 & 29.26$\pm$1.19                             \\ 
\hline

\multirow{2}{*}{FedBN}      & BN                   & 95.74$\pm$0.40 & 86.92$\pm$3.48 & 97.99$\pm$0.57 & 85.58$\pm$1.65 & 79.19$\pm$2.80 & \textbf{89.08$\pm$1.10}                   & 59.39$\pm$0.47 & 40.73$\pm$4.78 & 64.86$\pm$2.52 & 35.82$\pm$1.69 & 17.94$\pm$1.09 & \textbf{43.75$\pm$1.08}                    \\
                            & GN                   & 91.59$\pm$0.19 & 87.02$\pm$0.26 & 95.33$\pm$0.36 & 85.14$\pm$0.76 & 80.25$\pm$2.79 & 87.86$\pm$0.54                            & 39.35$\pm$1.22 & 36.64$\pm$2.07 & 41.06$\pm$3.96 & 31.99$\pm$1.30 & 22.80$\pm$1.15 & 34.37$\pm$1.49                             \\ 
\hline
\multirow{2}{*}{FedAdam}    & BN                   & 64.50$\pm$3.55 & 65.53$\pm$2.89 & 68.50$\pm$2.40 & 64.64$\pm$2.24 & 71.46$\pm$9.06 & 66.93$\pm$2.80                            & 13.70$\pm$0.39 & 17.88$\pm$1.26 & 14.74$\pm$0.16 & 18.17$\pm$0.84 & 17.55$\pm$1.03 & 16.41$\pm$0.66                             \\
                            & GN                   & 91.97$\pm$0.26 & 80.63$\pm$1.83 & 95.22$\pm$0.55 & 70.65$\pm$2.14 & 75.77$\pm$1.10 & \textbf{82.85$\pm$0.32}                   & 40.34$\pm$6.00 & 29.47$\pm$1.30 & 42.68$\pm$2.69 & 24.39$\pm$1.13 & 20.32$\pm$1.16 & \textbf{31.44$\pm$2.29}                    \\ 
\hline
\multirow{2}{*}{FedAdagrad} & BN                   & 62.42$\pm$3.43 & 62.89$\pm$3.33 & 66.54$\pm$1.42 & 63.94$\pm$2.29 & 70.84$\pm$8.58 & 65.33$\pm$1.80                            & 13.78$\pm$0.61 & 18.34$\pm$2.35 & 14.54$\pm$0.14 & 18.45$\pm$1.19 & 17.75$\pm$0.48 & 16.57$\pm$0.86                             \\
                            & GN                   & 92.26$\pm$0.89 & 84.28$\pm$0.52 & 95.77$\pm$0.13 & 70.99$\pm$0.69 & 76.22$\pm$1.20 & \textbf{83.90$\pm$0.11}                   & 42.38$\pm$6.66 & 32.73$\pm$6.66 & 45.59$\pm$3.03 & 23.02$\pm$0.69 & 20.84$\pm$2.24 & \textbf{32.91$\pm$1.72}                    \\ 
\hline
\multirow{2}{*}{FedYoGi}    & BN                   & 62.29$\pm$5.18 & 65.61$\pm$2.92 & 65.83$\pm$3.78 & 64.26$\pm$2.16 & 72.32$\pm$7.92 & 66.06$\pm$2.47                            & 13.87$\pm$0.81 & 19.07$\pm$0.95 & 14.89$\pm$0.53 & 18.63$\pm$0.64 & 18.22$\pm$0.26 & 16.94$\pm$0.45                             \\
                            & GN                   & 91.10$\pm$1.49 & 79.96$\pm$0.95 & 95.19$\pm$0.69 & 68.20$\pm$1.90 & 75.17$\pm$1.42 & \textbf{81.92$\pm$0.76}                   & 38.65$\pm$4.30 & 29.51$\pm$1.99 & 40.54$\pm$8.61 & 22.56$\pm$0.71 & 21.14$\pm$2.13 & \textbf{30.48$\pm$2.99}                    \\ 
\hline
\multirow{2}{*}{FedDyn}     & BN                   & 80.75$\pm$0.85   & 83.45$\pm$0.62   & 91.98$\pm$0.42   & 72.62$\pm$1.20   & 85.61$\pm$0.92   & 82.88$\pm$0.06  & 20.73$\pm$0.53   & 24.14$\pm$0.27   & 21.20$\pm$0.29   & 23.55$\pm$0.46   & 21.27$\pm$0.22   & 22.18$\pm$0.09                               \\
                            & GN                   & 88.42$\pm$0.14 & 86.35$\pm$1.12 & 95.47$\pm$0.18 & 82.73$\pm$0.98 & 82.19$\pm$1.07 & \textbf{87.03$\pm$0.21}                   & 28.61$\pm$1.11 & 31.05$\pm$1.18 & 41.38$\pm$3.58 & 31.93$\pm$0.71 & 20.60$\pm$1.06 & \textbf{30.71$\pm$1.04}                    \\ 
\hline
\multirow{2}{*}{\mname}     & BN                   & 95.93$\pm$0.30 & 88.81$\pm$1.86 & 97.64$\pm$0.92 & 84.36$\pm$1.70 & 84.51$\pm$3.62 & \textbf{\textcolor{blue}{90.25$\pm$1.27}} & 59.42$\pm$0.53 & 38.77$\pm$2.75 & 69.17$\pm$4.15 & 34.81$\pm$1.42 & 17.85$\pm$1.11 & \textbf{\textcolor{blue}{44.00$\pm$1.30}}  \\
                            & GN                   & 91.60$\pm$0.15 & 87.61$\pm$0.87 & 94.49$\pm$0.93 & 84.39$\pm$0.30 & 78.35$\pm$3.03 & 87.29$\pm$0.88                            & 36.71$\pm$0.78 & 35.39$\pm$3.36 & 41.17$\pm$0.03 & 31.82$\pm$1.18 & 19.80$\pm$0.61 & 32.98$\pm$1.03                             \\
\bottomrule
\end{tabular}
}
\vspace{-3mm}
\end{table*}

\vspace{-3mm}
\subsection{Statistical analysis}
\vspace{-1mm}
We conducted extensive experiments to provide comprehensive benchmarks. For example, for eICU dataset, there are 6 tasks, 8 algorithms, and 2 normalization techniques for each method. Also, we trained each model in the setting of 5, 10, 20, and 30 clients. For each experiment, each client should train the model respectively and the model should be aggregated. We had to go through this communication round 100, 20, or 10 times depending on the number of local training epochs. The same thing happened for the skin cancer dataset and ECG dataset. Due to this extensive process, we took 3 random seeds for each experiment. Each experiment is independent, but as there are three samples for each, we used the Wilcoxon rank-sum test (i.e. Mann–Whitney U test)\citep{Mann1947OnAT} to evaluate the statistical significance of the difference between each algorithm. The Wilcoxon rank-sum test is the non-parametric version of the two-sample t-test. We considered the result is statistically significant if a p-value is less than 0.05.

\vspace{-4.5mm}
\section{Results}
\vspace{-1mm}
\subsection{Experimental results for each dataset}
\label{sec:exp_results}
\subsubsection{eICU dataset}
\vspace{-1mm}
\label{eicu_experiment}
We present the results of the FL methods for six clinical prediction tasks on the eICU dataset described in \sectionref{sec:exp_setup_ehr}.
We present the experiment results using the data from the five largest clients with hospital IDs $73, 264, 420, 243, \text{and } 458$ in Table \ref{tab:eicu_5_clients_average}.

    

As shown in Table \ref{tab:eicu_5_clients_average} and Figure \ref{fig:eicu_stat}, there is no clear winner in the eICU dataset that outperforms all other algorithms across all tasks. The tendency varies even within the same discharge prediction task according to the time window (disch-24, disch-48). Therefore, we averaged AUROC of all six tasks to give better suggestions to the community.

We observe that most FL methods perform better when trained with GN instead of LN. We also observe that the average performance of our approach \mname trained using GN across all tasks is better than that of existing FL methods and our approach trained using LN.
We report the result of the Wilcoxon rank-sum test that shows the hybrid method never loses significantly on Figure~\ref{fig:eicu_stat}.  
\mname also outperforms all FL methods in terms of AUPRC, which can be viewed on Appendix Table~\ref{tab:eicu_5_clients_apurc}.


\vspace{-1mm}
\subsubsection{Skin cancer image dataset}
\label{sec:skin_cancer_experiment}
We present the results of FL methods on the skin cancer image dataset for the 8-way classification task with five clients for both BN and GN. Table \ref{tab:skin_cancer_average} reveals that training with BN is better than GN in the FedAvg, FedProx, FedBN, and \mname when comparing clients' average AUROC and AUPRC results. In contrast, GN is better than BN in FedAdam, FedAdagrad, FedYoGi and FedDyn.
\mname yields the overall best performance when trained with BN and GN. When using GN, FedBN
also shows good performance when compared to other methods. 
\begin{figure}[h!]
\floatconts
    {fig:skin_ecg_stat}
    {\caption{Results of the statistical test with the skin cancer and ECG datasets. Two red astericks(**) show that FedPxn wins significantly(i.e. p-value is less than 0.05). For the skin cancer datasets, FedPxN always wins significantly except one case. For the ECG datasets, FedPxN always wins significantly. \vspace{-8mm}}}
    {
        \centering
        \subfigure[skin cancer datasets]{\label{fig:stat-skin}
          \includegraphics[width=0.87\columnwidth]{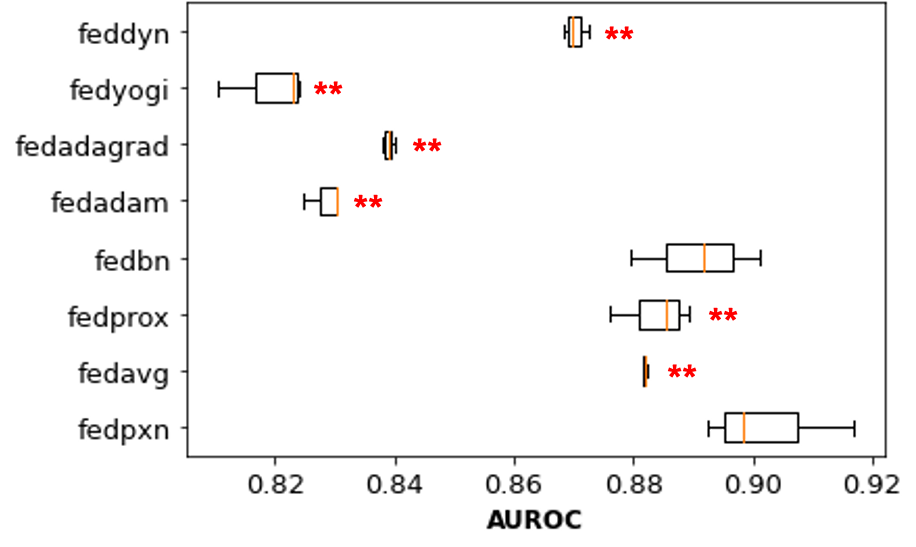}}
        \vspace{4mm}
        \subfigure[ECG datasets]{\label{fig:stat-ecg}
          \includegraphics[width=0.83\columnwidth]{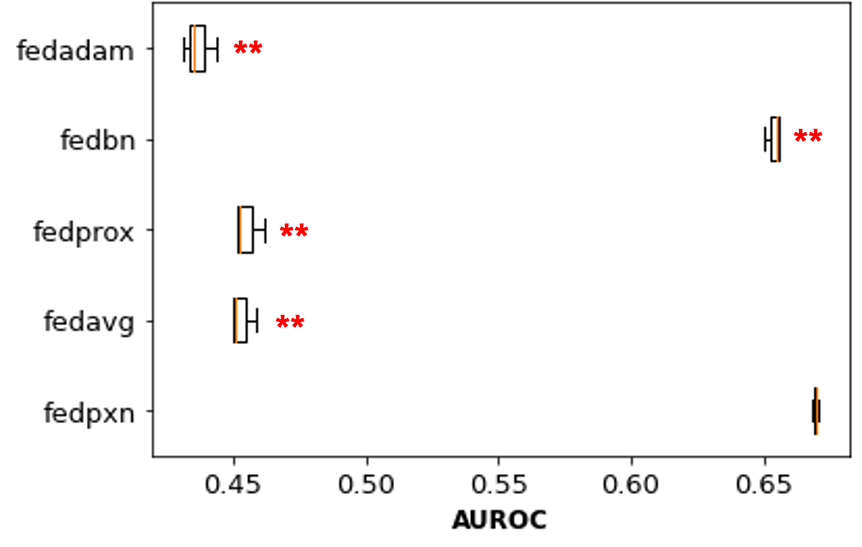}}
      \vspace{-9mm}
    }
\end{figure}
According to Figure~\ref{fig:stat-skin}, FedPxN never loses compared to each method with its better normalization technique and always wins significantly except FedBN.

Surprisingly, training was extremely unstable when FedDyn was trained using BN. 
Specifically, a $NaN$ value was found in the BN parameter during training.
To address the problem, we tried 1) extensive hyper-parameter tuning, 2) allowing BN's rescaling parameters to be aggregated \citep{andreux2020siloed} 3) allowing none of BN parameters to be aggregated \citep{li2021fedbn}, but none of the approaches helped improve the performance of FedDyn.

\subsubsection{ECG dataset}
\label{ecg_experiment}
We present the results of the FL methods for the 26 multi-label prediction task on ECG signals in Table \ref{tab:ecg_result}. All the results indicate that training with BN is more effective in FedAvg, FedProx, FedBN and \mname whereas training with GN is better for FedAdam. 
Our proposed approach, \mname, shows overall the best performance based on the average CinC score when trained with BN or GN. \mname always wins significantly as shown in Figure~\ref{fig:stat-ecg}.

FedDyn failed to train with either BN or GN (similar to \sectionref{sec:skin_cancer_experiment}). To address this failure in training, we tried various modifications similar to those stated in \sectionref{sec:skin_cancer_experiment}. However, we still encountered failure in training. Therefore, we do not report the results of FedDyn.

\begin{table}[h!]
\centering
\caption{CinC scores for the ECG dataset. We indicate the best normalization strategy for each method in bold. The highest average CinC is indicated in blue. }
\vspace{-1mm}
\label{tab:ecg_result}
\resizebox{\columnwidth}{!}{
\begin{tabular}{c|cccccc|c} 
\toprule
\multicolumn{1}{c}{}     &                        & Shaoxing     & CPSC         & Ga           & Ningbo       & PTBXL        & \textbf{Average}                      \\ 
\hline
\multirow{2}{*}{FedAvg}  & BN                     & 65.8$\pm$1.3 & 45.1$\pm$1.0 & 39.0$\pm$0.3 & 63.5$\pm$1.0 & 13.1$\pm$1.4 & \textbf{45.3$\pm$0.4}                 \\
                         & GN                     & 65.4$\pm$1.0 & 46.1$\pm$0.4 & 37.6$\pm$0.8 & 60.7$\pm$0.2 & 11.1$\pm$1.1 & 44.17$\pm$0.6                         \\ 
\hline
\multirow{2}{*}{FedProx} & BN                     & 66.9$\pm$1.4 & 47.3$\pm$0.8 & 40.0$\pm$0.1 & 58.7$\pm$0.9 & 14.7$\pm$0.4 & \textbf{45.53$\pm$0.5}                \\
                         & GN                     & 65.3$\pm$0.6 & 46.2$\pm$0.1 & 37.1$\pm$0.1 & 59.4$\pm$1.0 & 10.4$\pm$0.8 & 43.68$\pm$0.1                         \\ 
\hline
\multirow{2}{*}{FedBN}   & \multicolumn{1}{l}{BN} & 78.9$\pm$0.7 & 72.5$\pm$0.8 & 46.5$\pm$0.5 & 76.7$\pm$0.3 & 52.2$\pm$1.9 & \textbf{65.37$\pm$0.3}                \\
                         & GN                     & 77.6$\pm$0.3 & 68.5$\pm$1.5 & 49.2$\pm$0.4 & 75.9$\pm$0.5 & 49.7$\pm$0.3 & 64.19$\pm$0.3                         \\ 
\hline
\multirow{2}{*}{FedAdam} & BN                     & 61.6$\pm$1.3     & 43.9$\pm$3.0     & 33.1$\pm$1.6     & 56.9$\pm$5.1     & 4.3$\pm$1.3      & 39.95$\pm$1.2                             \\
                         & GN                     & 64.3$\pm$1.1     & 44.7$\pm$0.5     & 37.4$\pm$1.0     & 60.5$\pm$1.2     & 11.5$\pm$0.9     & \textbf{43.67$\pm$0.6}                             \\ 
\hline
\multirow{2}{*}{\mname}   & BN                     & 78.0$\pm$1.0 & 72.7$\pm$0.3 & 51.7$\pm$1.5 & 76.4$\pm$0.5 & 55.8$\pm$1.1 & \textbf{\textcolor{blue}{66.93$\pm$0.1}}  \\
                         & GN                     & 78.4$\pm$0.6 & 70.0$\pm$1.1 & 48.7$\pm$0.7 & 75.9$\pm$0.5 & 51.8$\pm$0.8 & 64.95$\pm$0.4                         \\
\bottomrule
\end{tabular}
}
\vspace{-3mm}
\end{table}


\begin{figure*}[h!]
\centering
\stackunder[5pt]{\includegraphics[width=1.9\columnwidth]{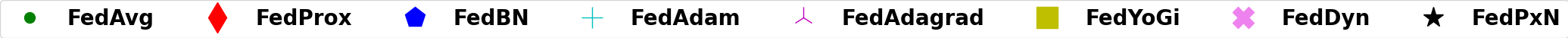}}{}
\stackunder[5pt]{\includegraphics[width=0.5\columnwidth]{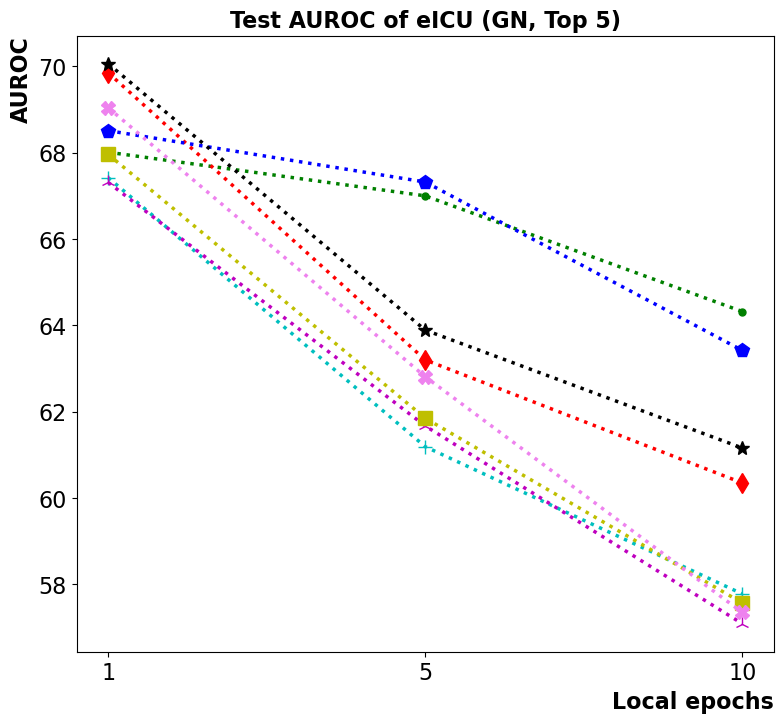}}{}
\hspace{-2.8mm}
\stackunder[5pt]{\includegraphics[width=0.5\columnwidth]{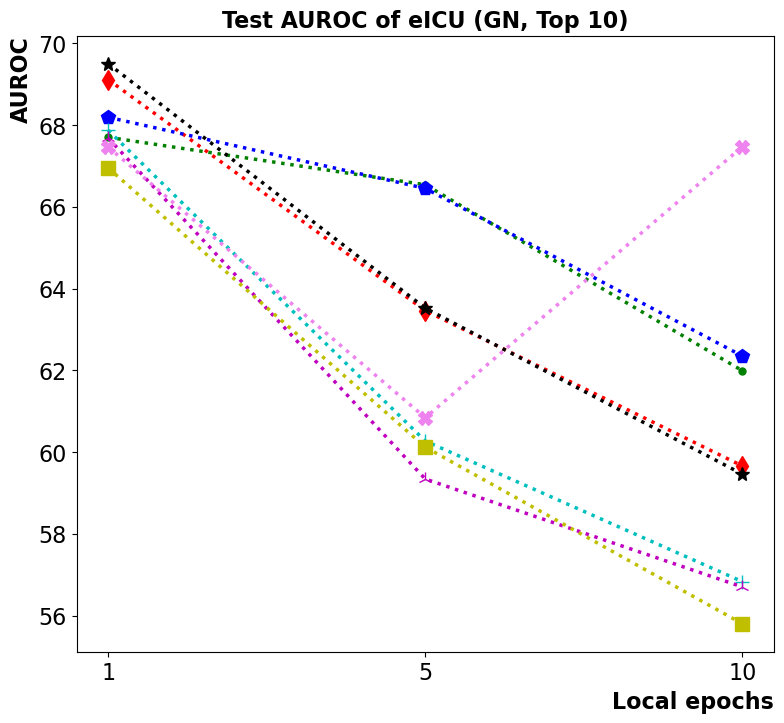}}{}
\hspace{-2.8mm}      
\stackunder[5pt]{\includegraphics[width=0.5\columnwidth]{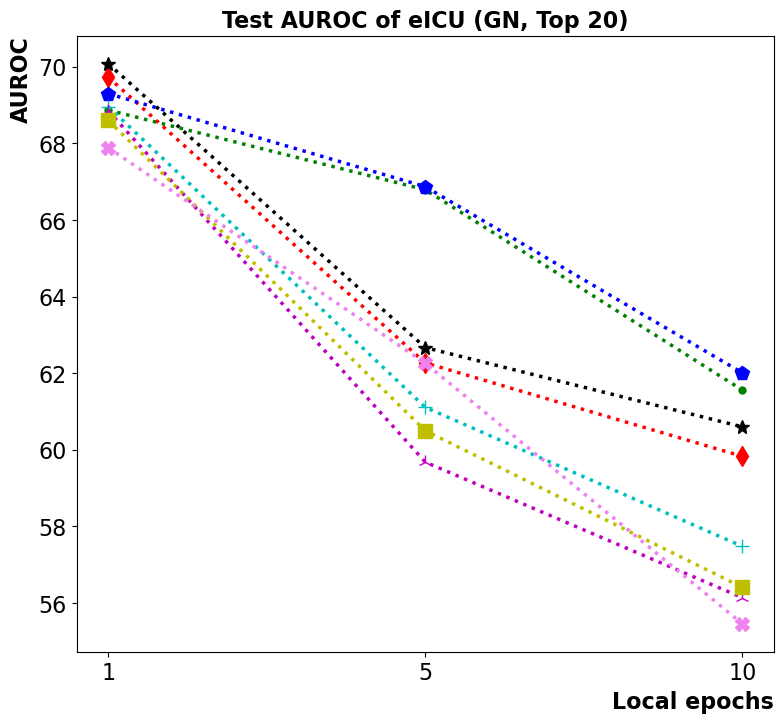}}{}
\hspace{-2.8mm}
\stackunder[5pt]{\includegraphics[width=0.5\columnwidth]{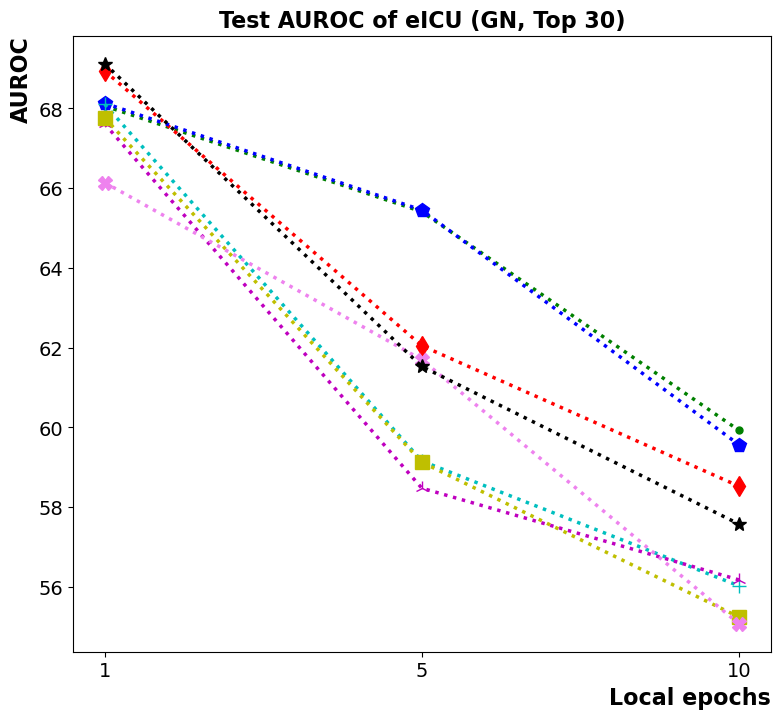}}{}
\stackunder[5pt]{\includegraphics[width=0.5\columnwidth]{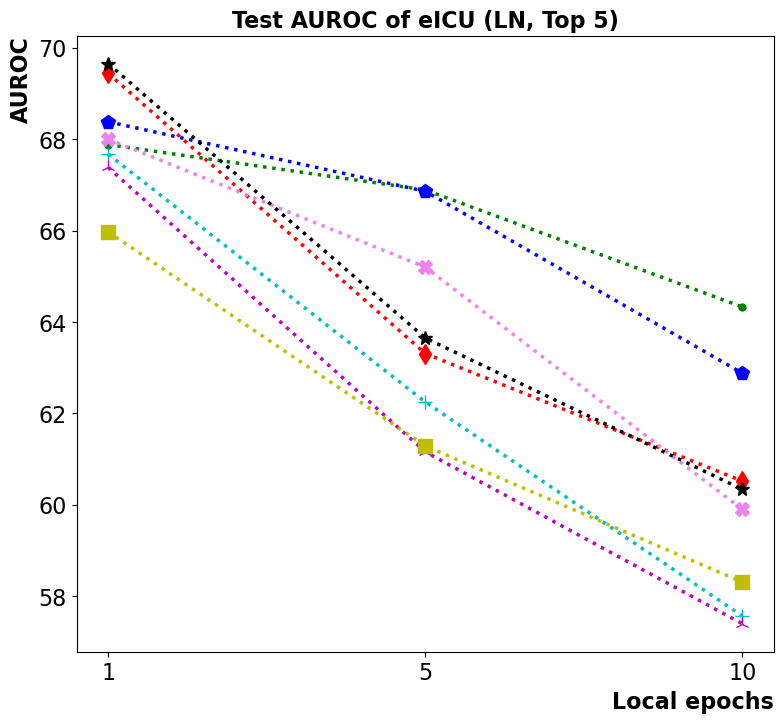}}{}
\hspace{-2.8mm}
\stackunder[5pt]{\includegraphics[width=0.5\columnwidth]{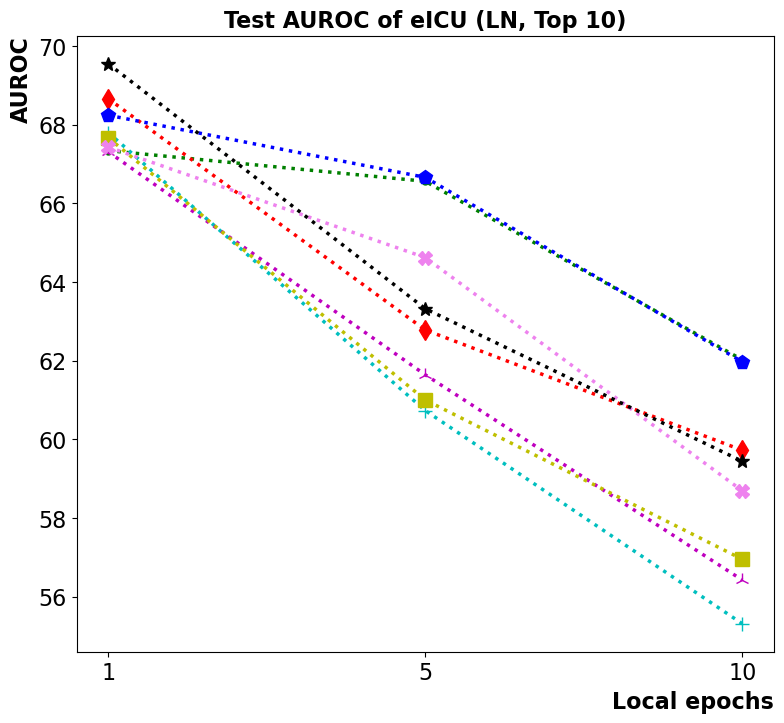}}{}
\hspace{-2.8mm}
\stackunder[5pt]{\includegraphics[width=0.5\columnwidth]{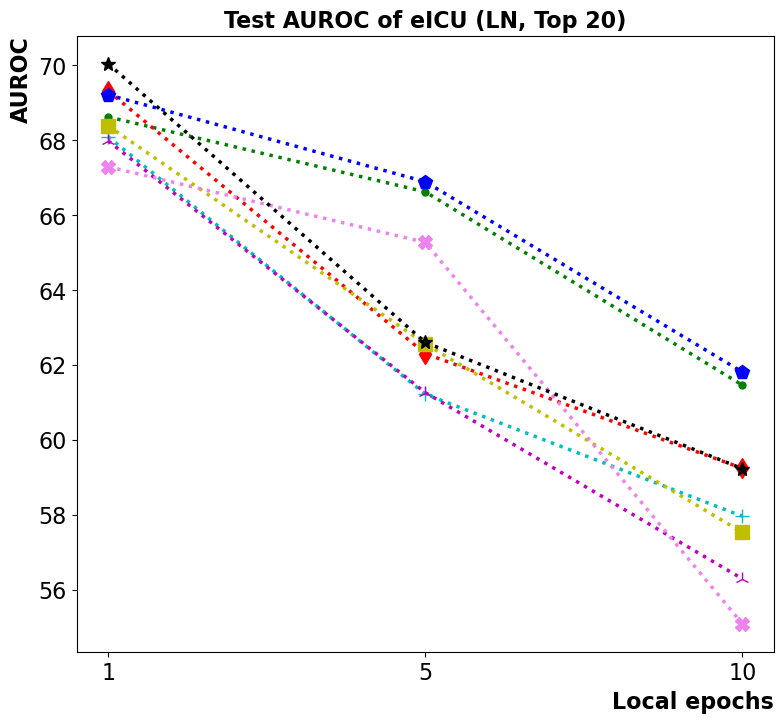}}{}
\hspace{-2.8mm}
\stackunder[5pt]{\includegraphics[width=0.5\columnwidth]{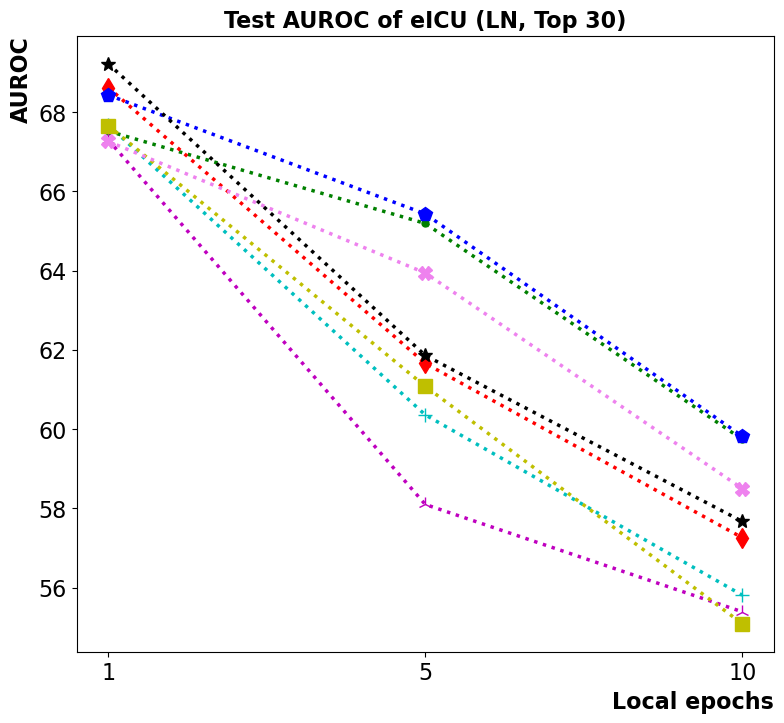}}{}
\vspace{-2.5mm}
\caption{Test AUROC results for the eICU dataset as the number of local training epochs increases.
 \vspace{-2mm}}
\label{fig:eicu_results}
\end{figure*}



\begin{figure*}[h!]
\centering
\vspace{2mm}
\stackunder[5pt]{\includegraphics[width=1.9\columnwidth]{images/eicu_legend.png}}{}
\stackunder[5pt]{\includegraphics[width=0.49\columnwidth]{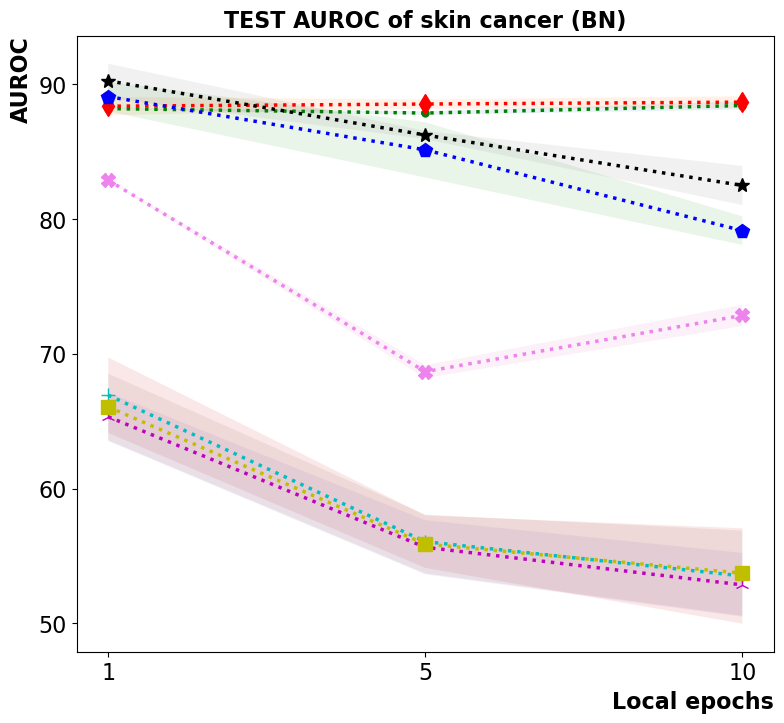}}{}
\hspace{-2.8mm}
\stackunder[5pt]{\includegraphics[width=0.49\columnwidth]{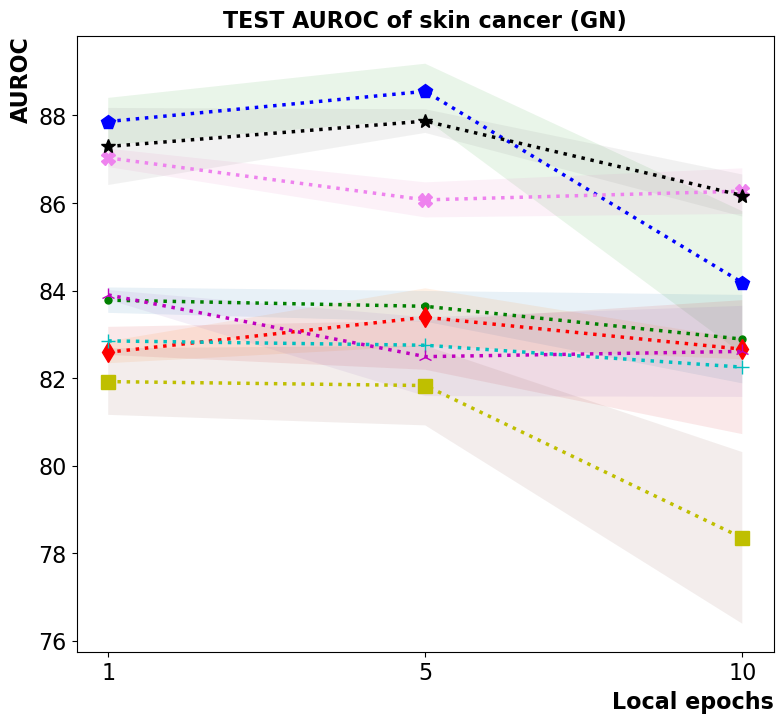}}{}
\hspace{-2.8mm}
\stackunder[5pt]{\includegraphics[width=0.49\columnwidth]{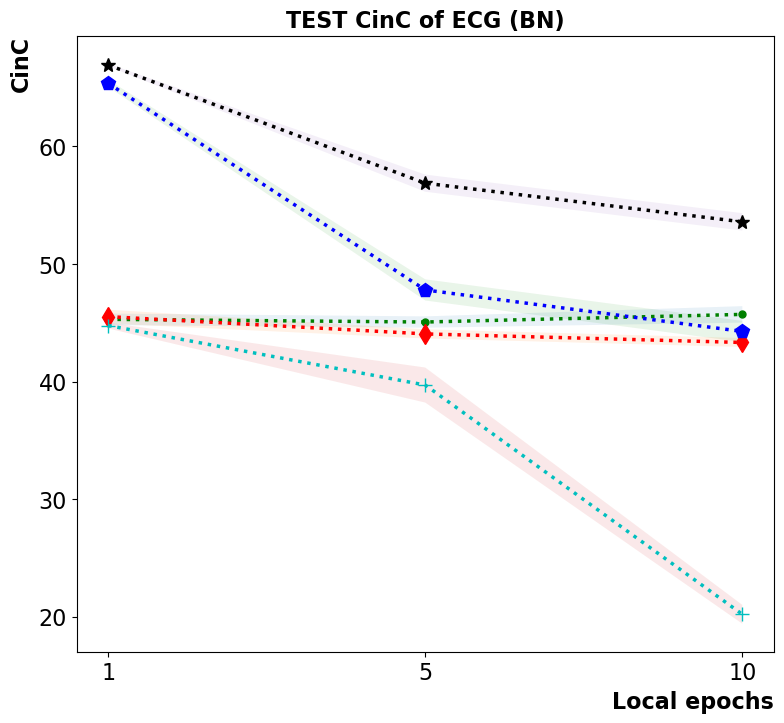}}{}
\hspace{-2.8mm}
\stackunder[5pt]{\includegraphics[width=0.49\columnwidth]{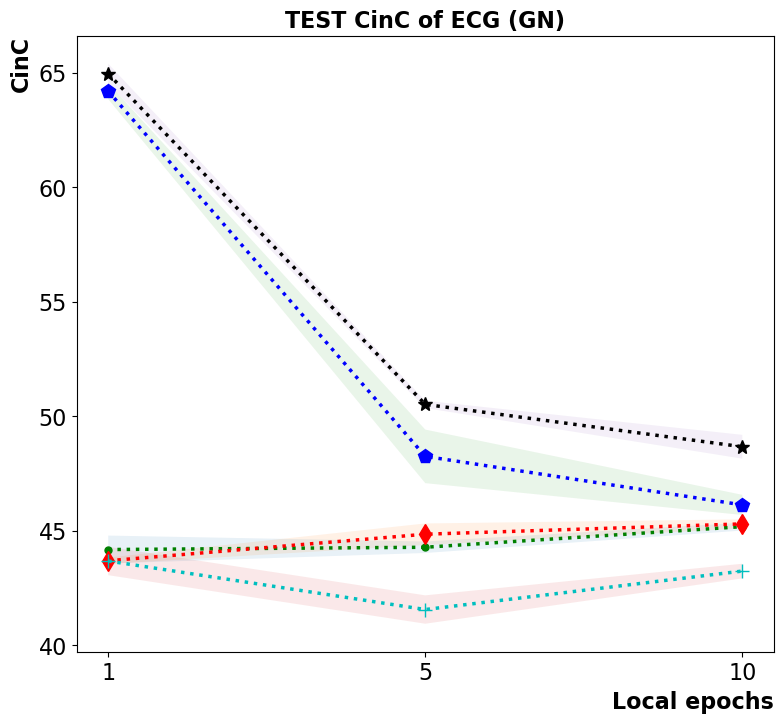}}{}
\vspace{-2.5mm}
\caption{Test AUROC and CinC results for the skin cancer image and ECG datasets measured as the number of local training epochs increases. \vspace{-2.5mm}}
\label{fig:skin_eicu_results}
\end{figure*}



\subsection{The number of local training epochs}
\label{ssec:local_epoch}
We investigated the effect of the number of local epochs on the performance of the FL algorithms.
We conducted experiments on the three datasets (eICU, skin cancer, and ECG) using $1$, $5$, and $10$ local epochs, while fixing the number of total training epochs.
The performance was averaged across three seeds.
Figure \ref{fig:eicu_results} presents the results on the eICU dataset, where the images in the top and bottom rows show the plots of models trained with GN and LN, respectively.
Each row contains four plots corresponding to models trained and tested on $5, 10, 20,$ and $30$ clients by AUROC score averaged across all six tasks.
Figure \figureref{fig:skin_eicu_results} presents the results of models trained with BN or GN and varying local epochs on the skin cancer and ECG dataset.

Figures \ref{fig:eicu_results} and \ref{fig:skin_eicu_results} reveal that the performance of most FL methods including \mname consistently decreases as the number of local epochs increases.
This demonstrates that FL methods perform well with fewer local epochs and more global updates.
Because the number of total training epochs is fixed, fewer local epochs indicate more communication rounds, which would require good internet connectivity and might result in increased bandwidth costs.
However, in medicine, a model's performance is the utmost priority, as it has a high impact on a patient's life and a wrong prediction could lead to fatal errors.
Therefore, although more communication rounds increase communication costs, a better performing FL model is preferable in medicine.

\subsection{Overall guidance from the results}
\label{ssec:guidance}
\vspace{-2mm}
The first step to apply FL is to decide the FL algorithm. \mname shows favorable performances for every dataset. For the structured data, there was no clear winner, but \mname is always included in the top 3 powerful methods, often with FedProx and FedDyn. For the image or signal data, \mname is a clear winner overall and FedBN follows. In addition, \mname has relatively less fluctuation with its performance, so it can be the first option to consider. FedAvg is the most basic and simple algorithm but never wins in our experiments. Unfortunately, FedOpt variants were poor with the image or signal dataset. 

The next step is to design a communication process. According to our results above, it is better to train local clients’ models with a single local epoch and increase the number of communication rounds instead. This guidance will be a good starting point for the hospitals or the IT companies to adopt FL. 

\section{Discussion}
We observe which method is cost-effective for optimal performance by measuring power consumption each FL method requires to achieve the best performance. As the number of clients increases in FL, more IT administrators and GPUs will be needed, and the related monetary costs will increase accordingly. So we note which methods are effective as the number of clients increases using the eICU database. Furthermore, we explain our limitations in this section.
\subsection{Power consumption}
We measured the power consumed while conducting the experiments using a single local epoch. We report the results for one seed value only for this experiment in \tableref{tab:power_consumption}. To understand the monetary costs of each method to achieve optimal performance, we conducted experiments using the normalization technique that showed better performance in \sectionref{sec:exp_results}. During training, we queried the NVIDIA System Management Interface \footnote{https://developer.nvidia.com/nvidia-system-management-interface} at regular intervals to measure the power consumption of GPU and averaged the measured values \citep{strubell2019energy}. Then, we multiplied the obtained average values by elapsed time.
The power consumption of all FL methods is comparable. The difference between max and min power consumption among all methods in each dataset is between 0.103-0.596 kWh. The number of hyperparameters varies depending on the FL method (see Appendix~\ref{appendix:hyperparameter}). So, the hyperparameter-tuning cost of FedAvg and FedBN is the cheapest considering the number of hyperparameters. Practically, FedBN is the cheapest technique for all tasks on the three datasets. The power consumption of \mname is 1.03-1.17 times more than that of FedBN in \tableref{tab:power_consumption}.

\begin{table}[h!]
\centering
\caption{Power consumption of each method for the tasks in the eICU, skin cancer images, and ECG datasets. In eICU, we measured total power consumption of six tasks with 30 clients.}
\vspace{-2mm}
\label{tab:power_consumption}
\resizebox{\columnwidth}{!}{
\begin{tabular}{lc|cccccccc} 
\toprule
\multicolumn{1}{c}{} &                         & FedAvg & FedProx & FedBN                            & \mname & FedAdam                          & FedAdagrad           & FedYoGi              & FedDyn                \\ 
\hline\hline
\multicolumn{10}{c}{\textbf{eICU}}                                                                                                                                                                                    \\ 
\hline\hline
\multicolumn{1}{c}{} &                         & GN     & GN      & LN                               & LN    & GN                               & GN                   & GN                   & LN                    \\ 
\hline
\multicolumn{1}{c}{} & elapsed time (h)        & 5.66   & 5.82    & 4.51                             & 4.51  & 4.81                             & 4.84                 & 5.14                 & 5.13                  \\
\multicolumn{1}{c}{} & Power consumption (kWh) & 0.542  & 0.560   & 0.510                            & 0.526 & \textbf{\textcolor{blue}{0.479}} & 0.484                & 0.518                & 0.582                 \\
\hline\hline
\multicolumn{10}{c}{\textbf{Skin cancer images}}                                                                                                                                                                      \\ 
\hline\hline
                     & \multicolumn{1}{l|}{}   & BN     & BN      & BN                               & BN    & GN                               & GN                   & GN                   & GN                    \\ 
\hline
\multicolumn{1}{c}{} & elapsed time (h)        & 10.53  & 10.62   & 10.29                            & 10.27 & 10.56                            & 10.67                & 10.61                & 9.49                  \\
                     & Power consumption (kWh) & 3.034  & 3.280   & \textbf{\textcolor{blue}{2.989}} & 3.266 & 3.046                            & 3.097                & 3.100                & 3.072                 \\
\hline\hline
\multicolumn{10}{c}{\textbf{ECG}}                                                                                                                                                                                     \\ 
\hline\hline
                     & \multicolumn{1}{l|}{}   & BN     & BN      & BN                               & BN    & GN                               &                      &                      &                       \\ 
\hline
                     & elapsed time (h)        & 9.14   & 11.49   & 9.15                             & 10.71 & 9.58                             & \multicolumn{1}{l}{} & \multicolumn{1}{l}{} & \multicolumn{1}{l}{}  \\
                     & Power consumption (kWh) & 2.955  & 3.484   & \textbf{\textcolor{blue}{2.888}}                            & 3.401 & 3.016 & \multicolumn{1}{l}{} & \multicolumn{1}{l}{} & \multicolumn{1}{l}{}  \\
\bottomrule
\end{tabular}
}
\vspace{-4mm}
\end{table}

\subsection{Administrators and GPUs}
\subsubsection{Training with more clients}
We performed experiments in which we trained the FL methods with 10, 20, 30 largest clients and reported the average AUROC of all six tasks using the eICU in Appendix \tableref{tab:eicu_total_average_auroc}. GN outperforms LN in most FL methods trained with $5-30$ clients. In addition, we observe that \mname yields the best AUROC in each setting (5, 10, 20, 30 clients) despite the similar monetary costs requirement of different FL methods.
\subsubsection{Testing on the top-30 client dataset with varying training clients}
\begin{figure}[h!]
    \centering
    {
        \stackunder[5pt]{ \includegraphics[width=0.9\columnwidth]{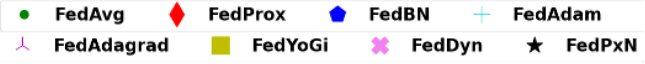} }{}
        \stackunder[5pt]{ \includegraphics[width=0.48\columnwidth]{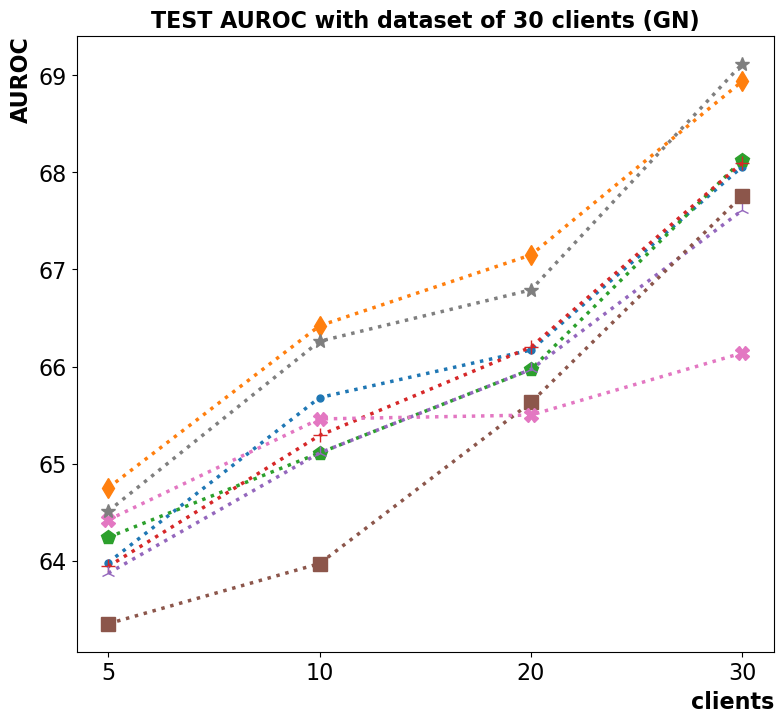} }{}
        \hspace{-5mm}
        \stackunder[5pt]{ \includegraphics[width=0.48\columnwidth]{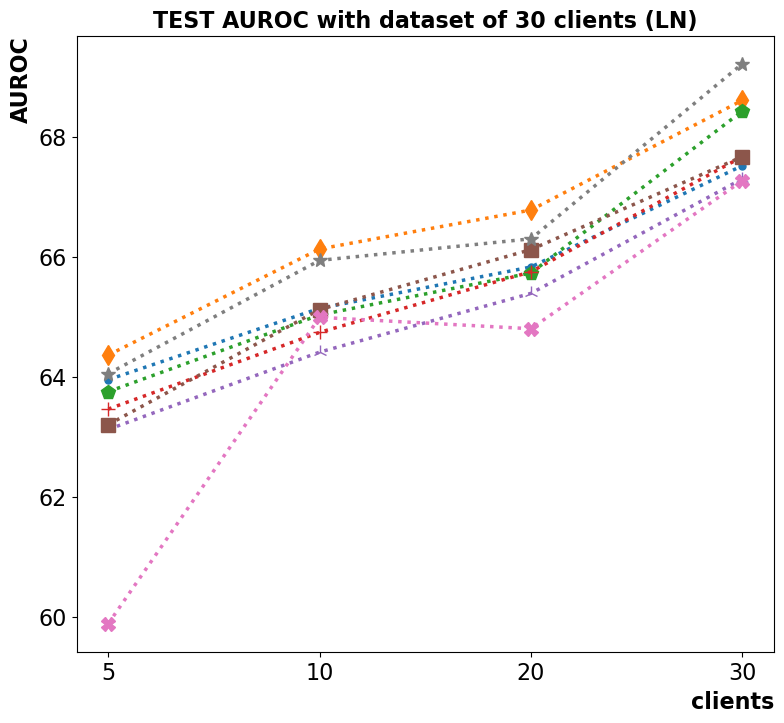} }{}
}
\vspace{-3.5mm}
\caption{Test AUROC results for the five largest clients and all 30 clients measured using the models trained on varying numbers of clients.}
\label{fig:eicu_clients_5_30}
\vspace{-2.5mm}
\end{figure}

We evaluated all FL methods on the test set from the top 30 clients, while varying the number of clients the models were trained on. The performance is measured by the average AUROC of all six tasks. As depicted in Figure~\ref{fig:eicu_clients_5_30}, the AUROC results of all methods except FedDyn consistently increase when both LN and GN are used.

This demonstrates that training the model on more clients generalizes the model due to the availability of heterogeneous data and improves model performance. FedProx obtains the best performance with both GN and LN when the FL methods are trained on 20 or fewer clients. This implies that training with FedProx using fewer hospitals can provide better performance for hospitals not involved in training. However, there is only a marginal difference between the AUROC of FedProx and \mname. For example, the AUROC of FedProx and \mname are 67.15 and 66.79 when training with 20 clients’ data using GN. Also, \mname yields the optimal performance when trained on the data of all 30 clients.

\subsection{Transmitting cost}
For a fair comparison, we fixed the number of total communication rounds for each task, and therefore the transmitting fee was not within our scope. Yet, we assume that the transmitting fee is insignificant. Institutions such as hospitals usually pay the fee for the internet connection on a monthly or yearly basis. In our experimental settings, the model parameters were approximately 4.4MB, 16MB, and 32MB and the cost of exchanging model parameters 100-300 times is sufficiently covered by the amount paid. Therefore, there is no practical need to calculate the transmitting fee separately.\\
However, if the hospital uses the cloud service, it would be better to consider it carefully because the transmission cost may vary depending on the bandwidth, and increase proportionally with the model size. In that case, it would be an interesting future work to analyze the FL framework that guarantees generally good performance while minimizing the communication rounds by fast convergence.

\subsection{Limitations}
Although we provide comprehensive benchmarks on three medical modalities for the first time, there are some limitations. We conducted experiments across multiple GPU’s but on the same machine for a fair experimental setting. It would have been even more realistic if the experiments were conducted across independent test sites. 
Plus, eICU database is across independent hospitals, but they are still from the United States. Therefore, it does not fully reflect real-world multi-national heterogeneity. For medical images in our experimental setting, we tested FL methods for a specific cancer and task. The next step is to validate FL methods across different clinical tasks or cancer types (e.g., medical image segmentation, breast cancer prediction).

\section{Conclusion}
In this work, we provide practical benchmarks of various FL algorithms on the real-world heterogeneous datasets in the clinical domain with three modalities: structured, image, and signal data. We find out that a hybrid algorithm, \mname, that introduces the regularization term from FedProx to the FedBN framework is simple but effective. It has a slightly higher power consumption than the most economical method (FedBN), but mostly outperforms the other methods and never loses significantly. We expect users to refer to these experimental results to design their FL settings and to select a suitable FL algorithm considering their clinical tasks and obtain stronger performance with lower costs. 

\vspace{3mm}

\paragraph*{Institutional Review Board (IRB)}
This research does not require IRB approval.


\acks{
This work was supported by the KAIST Key Research Institute (Interdisciplinary Research Group) Project, Institute of Information \& Communications Technology Planning \& Evaluation (IITP) grant (No.2019-0-00075), and the Korea Health Industry Development Institute (KHIDI) grant (No.HR21C0198, No.HI22C1518), funded by the Korea government (MSIT, MOHW).
}

\clearpage

\bibliography{jmlr-sample}

\clearpage
\appendix
\section{FL methods}
\label{appendix:explain_methods}
In this section, we describe the FL methods that are evaluated in the experiments.
\vspace{-4mm}
\subsection{FedAvg}
\vspace{-4mm}
\setlength{\textfloatsep}{8pt}
\begin{algorithm}[h!]
\small
\floatconts
{algo:fedavg}
{\caption{Federated Averaging}}
{
\SetAlgoLined
\SetKwInput{KwInput}{Input}
\SetKwInput{KwOutput}{Output}
\KwInput{number of clients $K$, number of communication rounds $T$, number of local epochs $E$, Data $D$ := $(D_{1}, D_{2}, ..., D_{K})$, learning rate $\eta$}
\KwOutput{model parameter $w_{T}$}

\textbf{Server executes:} \\
\Indp
initialize model parameters $w_{0}$ \\
\Indp
\For{$t = 0,..., T - 1$}{
  
  \For{each client $k \in K$ } {
    $w_{t,k}$ $\leftarrow$ $w_{t}$ \\
    $w_{t,k}$ $\leftarrow$ LocalTraining($k$, $w_{t,k}$, $D_{k}$) \\
  }
  
    $w_{t+1}$ $\leftarrow$ $\sum_{k=1}^{K} \frac{n_k}{n} $ $w_{t,k}$ \\ 

}

\Indm 
\Indm 
\textbf{LocalTraining($k$, $w_{t,k}$, $D_{k}$):} \\
\Indp
\For{$e=0,...,E-1$}{
    \For{ batch b $\leftarrow$ (x, y) of $D_{k}$ } {
    $w_{t,k}$ $\leftarrow$ $w_{t,k}$ - $\eta\nabla F_k(w_{t,k}; b)$
}
}
\textbf{return} $w_{t,k}$
}
\vspace{-3mm}
\end{algorithm}

\vspace{-3mm}
\subsection{FedOpt}
FedOpt \citep{reddi2020adaptive} applies adaptive optimization to model the aggregation stage in FedAvg. First, the pseudo gradient $\Delta_{t,k} := w_{t,k} - w_{t}$ of each client $k$ is calculated after local training at each round $t$. Second, it calculates $\Delta_{t}$ by averaging each pseudo gradient $\Delta_{t,k}$. Third, the momentum $m_{t} \leftarrow \beta_{1}m_{t-1} + (1-\beta_{1})\Delta_{t}$ is calculated. Then, $v_{t}$ is calculated using different adaptation techniques (FedAdam, FedAdagrad, FedYoGi) as follows:
\vspace{-1.5mm}
\begin{equation}
\small
  \begin{aligned}
    & \text{FedAdam} \\
    & v_{t} \leftarrow \beta_{2}v_{t-1} + (1-\beta_{2})\Delta_{t}^2, \\
    & \text{FedAdagrad} \\
    & v_{t} \leftarrow v_{t-1} + \Delta_{t}^2, \\
    & \text{FedYoGi} \\
    & v_{t} \leftarrow v_{t-1} - (1-\beta_{2})\Delta_{t}^2sign(v_{t-1} - \Delta_{t}^2), \\
  \end{aligned}
  \label{eq:adaptation}
\end{equation}
where $\beta_{1}$, $\beta_{2}$ are the hyperparameters for adaptation. Finally, a global model is updated using $m_{t}$ and $v_{t}$ as follows:

\begin{equation}
\small
  \begin{aligned}
    & w_{t+1} \leftarrow w_{t} + \eta_{g}\frac{m_{t}}{\sqrt{v_{t}}+\gamma} &&,
  \end{aligned}
  \label{eq:aggregation_fedopt}
\end{equation}
where $\eta_{g}$ is the server learning rate and $\gamma$ represents the degree of adaptivity for each algorithm. 

\vspace{-1mm}
\subsection{FedPxN}
\begin{algorithm}[h!]
\small
\floatconts
{algo:fedpxn}
{\caption{Federated learning with Proximal regularization eXcept local Normalization (\mname)}}
{
\SetAlgoLined
\SetKwInput{KwInput}{Notation}
\KwInput{number of clients $K$, number of communication rounds $T$, number of local epochs $E$, Data $D$ := $(D_{1}, D_{2}, ..., D_{K})$, learning rate $\eta$, normalization layers $norm$}

\textbf{Server executes:} \\
\Indp
initialize model parameters $w_{0}$ \\
\Indp
\For{$t = 0,..., T - 1$}{
  \For{each client $k \in K$} {
    $w_{t,k \setminus norm}$ $\leftarrow$ $w_{t \setminus norm}$ \\
    $w_{t,k}$ $\leftarrow$ LocalTraining($k$, $w_{t,k}$, $D_{k}$) \\
  }
$w_{t+1 \setminus norm}$ $\leftarrow$ $\sum_{k=1}^{K} \frac{n_k}{n} $ $w_{t,k \setminus norm}$ \\ 
}
\Indm 
\Indm 
\textbf{LocalTraining($k$, $w_{t,k}$, $D_{k}$):} \\
\Indp
\For{$e=0,...,E-1$}{
    \For{ batch b $\leftarrow$ (x, y) of $D_{k}$ } {
    $R$ = $\| w_{t,k \setminus norm} - w_{t \setminus norm} \|^2$ \\
    $w_{t,k}$ $\approx$ $\underset{w}{\arg\min}$ $F_k(w;b)$ + $\frac{\mu}{2}$$R$
}
}
\textbf{return} $w_{t,k}$
}
\end{algorithm}

\vspace{-1mm}
\subsection{FedDyn}
In FedDyn \citep{acar2021federated}, the local model $w_{t,k}$ is updated by adding a penalized risk function to the objective function $F_{k}$ of FedAvg during local training. The risk objective of each client is dynamically updated using both local and global model as:
\begin{equation}
\small
  \begin{aligned}
    \hat{F}_{k}(w_{t,k};b) &=F_{k}(w_{t,k};b)-\Delta F_{k}(w_{t-1,k};b){w_{t,k}} \\
    &+ \frac{\alpha}{2}\| w_{t,k} - w_{t-1} \|^{2},
  \end{aligned}
\end{equation}
where $\alpha$ is the hyperparameter controlling the degree of regularization. Theoretically, local models converge to the global model if they converge in local training of FedDyn.

\section{Details of Experimental Setup}
\label{ex:details}
For each dataset, the total number of training epochs was set to $100, 300$ and $200$ respectively, which was sufficient for all models using all FL methods to converge.
After training, among model weights from all communication rounds $w_1, \ldots w_T$, the $w_t$ that showed the best average validation performance across all clients is chosen as the final model weight. Then we use the final model weight to calculate the average test performance across all clients.
We repeated all experiments three times and report the mean test performance.

\paragraph{eICU database}
\label{appendix:eicu_info}
The benchmark dataset in \cite{mcdermott2021comprehensive} contains labs, vitals, and demographic information.
The labs and vitals were measured for at least $5$\% of all observed time-points. 
The benchmark dataset contains $71477$ ICU stays across $59$ hospitals, where ICU stays range between 540 and 4008 for each hospital.
To conduct experiments in the FL setting, we used the $5$, $10$, $20$, and $30$ hospitals with the most ICU stays (with total of $14962, 25198, 39501, \text{and } 50434$ ICU stays) and performed the six clinical prediction tasks :
\begin{itemize} 
\item \textbf{Mortality prediction} (mort\_24h, mort\_48h)  \\
This task aims to predict whether the recorded time of death is within $24/48$ hours.\\
\textit{Input:} first $24$ hours of data;\\
\textit{Type:} binary classification. 
\item \textbf{Length-of-stay prediction} (LOS) \\
The LOS task aims to predict whether the patient's total stay is more than three days.\\
\textit{Input:} first $24$ hours of data; \\
\textit{Type:} binary classification. 
\item \textbf{Discharge prediction} (disch\_24h, disch\_48h) \\
This task predicts whether the patient is discharged within the next $24/48$ hours. If the patient is discharged, this task further aims to predict the next place of the patient (e.g., a skilled nursing facility or home).
\\
\textit{Input:} first $24$ hours of data; \\
\textit{Type:} 10-way classification.
\item \textbf{Final acuity prediction} (Acuity) \\ 
This task aims to predict whether a patient dies or is discharged. If the patient dies, it also predicts when the patient dies (e.g., in ICU or in hospital). If the patient is discharged, this task also aims to predict the next place of the patient.\\
\textit{Input:} first $24$ hours of data; \\
\textit{Type:} 10-way classification.
\end{itemize}
When using the five clients with the largest samples, the label distribution of six prediction tasks is described in Appendix table \ref{tab:eicu_label_distribution}. We used a $2$-layer Transformer encoder model \citep{vaswani2017attention} followed by two fully connected (FC) layers, similar to \citep{song2018attend}. We also applied Layer Normalization (LN) between the FC layers. We conducted additional experiments in which we replaced the LN used in the Transformer encoder and FC layers with Group Normalization (GN). 
We randomly split the ICU stays of each client using the ratio of 7:1.5:1.5 to form training, validation, and test sets.
To train the models for binary classification, and 10-way classification tasks, we used the binary cross-entropy loss and cross-entropy loss, respectively.
We used the Adam optimizer \citep{kingma2014adam}, a batch size of $256$, a single local epoch, $100$ communication rounds, and $100$ total training epochs in all of the tasks. 
We also tested varying combinations of the number of local epochs and communication rounds, such as 5 \& 20 and 10 \& 10, but 1 \& 100 generally gave the best performance for all FL methods in all tasks. 

\begin{table*}[htbp]
\centering
\caption{Label information of six tasks in eICU when using five clients with the largest samples (hospital id : 73, 264, 420, 243, 458). $NaN$ is unlabled data. We used only labeled data in all experiments. }
\vspace{-3mm}
\label{tab:eicu_label_distribution}
\resizebox{\textwidth }{!}{
\begin{tabular}{c|c|c|cccccccccc|c} 
\toprule
Task                        & Hospital ID & NaN & 0    & 1    & 2   & 3   & 4   & 5   & 6  & 7  & 8   & 9   & Total  \\ 
\hline\hline
\multirow{5}{*}{mort\_24h}  & 73          & 6   & 3978 & 24   & 0   & 0   & 0   & 0   & 0  & 0  & 0   & 0   & 4008   \\
                            & 264         & 6   & 3438 & 51   & 0   & 0   & 0   & 0   & 0  & 0  & 0   & 0   & 3495   \\
                            & 420         & 9   & 2754 & 55   & 0   & 0   & 0   & 0   & 0  & 0  & 0   & 0   & 2818   \\
                            & 243         & 3   & 2343 & 35   & 0   & 0   & 0   & 0   & 0  & 0  & 0   & 0   & 2381   \\
                            & 458         & 3   & 2218 & 39   & 0   & 0   & 0   & 0   & 0  & 0  & 0   & 0   & 2260   \\ 
\midrule
\multirow{5}{*}{mort\_48h}  & 73          & 11  & 3951 & 46   & 0   & 0   & 0   & 0   & 0  & 0  & 0   & 0   & 4008   \\
                            & 264         & 20  & 3384 & 91   & 0   & 0   & 0   & 0   & 0  & 0  & 0   & 0   & 3495   \\
                            & 420         & 21  & 2705 & 92   & 0   & 0   & 0   & 0   & 0  & 0  & 0   & 0   & 2818   \\
                            & 243         & 10  & 2322 & 49   & 0   & 0   & 0   & 0   & 0  & 0  & 0   & 0   & 2381   \\
                            & 458         & 8   & 2177 & 75   & 0   & 0   & 0   & 0   & 0  & 0  & 0   & 0   & 2260   \\ 
\midrule
\multirow{5}{*}{LOS}        & 73          & 0   & 2669 & 1339 & 0   & 0   & 0   & 0   & 0  & 0  & 0   & 0   & 4008   \\
                            & 264         & 0   & 2152 & 1343 & 0   & 0   & 0   & 0   & 0  & 0  & 0   & 0   & 3495   \\
                            & 420         & 0   & 1622 & 1196 & 0   & 0   & 0   & 0   & 0  & 0  & 0   & 0   & 2818   \\
                            & 243         & 0   & 1537 & 844  & 0   & 0   & 0   & 0   & 0  & 0  & 0   & 0   & 2381   \\
                            & 458         & 0   & 1512 & 748  & 0   & 0   & 0   & 0   & 0  & 0  & 0   & 0   & 2260   \\ 
\midrule
\multirow{5}{*}{disch\_24h} & 73          & 459 & 2132 & 1101 & 175 & 0   & 68  & 30  & 12 & 21 & 8   & 2   & 4008   \\
                            & 264         & 386 & 1979 & 794  & 215 & 0   & 44  & 26  & 10 & 26 & 0   & 15  & 3495   \\
                            & 420         & 404 & 1623 & 499  & 105 & 0   & 124 & 28  & 22 & 8  & 3   & 2   & 2818   \\
                            & 243         & 296 & 1259 & 667  & 89  & 0   & 0   & 37  & 5  & 25 & 0   & 3   & 2381   \\
                            & 458         & 345 & 1109 & 705  & 39  & 0   & 32  & 16  & 5  & 7  & 0   & 2   & 2260   \\ 
\midrule
\multirow{5}{*}{disch\_48h} & 73          & 792 & 1266 & 1456 & 263 & 0   & 103 & 59  & 23 & 29 & 15  & 2   & 4008   \\
                            & 264         & 555 & 1202 & 1163 & 373 & 0   & 76  & 44  & 15 & 45 & 0   & 22  & 3495   \\
                            & 420         & 540 & 1050 & 725  & 175 & 0   & 228 & 49  & 25 & 14 & 5   & 7   & 2818   \\
                            & 243         & 413 & 764  & 946  & 143 & 0   & 0   & 62  & 6  & 42 & 0   & 5   & 2381   \\
                            & 458         & 511 & 641  & 947  & 63  & 0   & 54  & 25  & 7  & 10 & 0   & 2   & 2260   \\ 
\midrule
\multirow{5}{*}{Acuity}     & 73          & 0   & 2642 & 609  & 255 & 138 & 47  & 46  & 28 & 2  & 45  & 196 & 4008   \\
                            & 264         & 0   & 1977 & 790  & 180 & 91  & 27  & 114 & 0  & 37 & 144 & 135 & 3495   \\
                            & 420         & 0   & 1379 & 392  & 555 & 79  & 42  & 30  & 11 & 12 & 220 & 98  & 2818   \\
                            & 243         & 0   & 1599 & 326  & 1   & 162 & 8   & 77  & 0  & 14 & 50  & 144 & 2381   \\
                            & 458         & 0   & 1674 & 128  & 138 & 54  & 11  & 18  & 0  & 6  & 176 & 55  & 2260   \\
\bottomrule
\end{tabular}
}
\end{table*}

\paragraph{Skin cancer dataset}
\label{appendix:skin_info}
Following \cite{cassidy2022analysis}, we used the ISIC19 \citep{codella2019skin} and HAM10000 \citep{tschandl2018ham10000} datasets, removing image samples that appear in both datasets from the ISIC19 dataset. Then, we split the HAM10000 \citep{tschandl2018ham10000} dataset into two datasets based on the source of the data sample. Hence, we formed three clients from the ISIC19 and HAM10000 datasets. The clients are from the Hospital Clinic de Barcelona (Barcelona), Medical University of Vienna, Austria (Vienna), and Queensland University, Australia (Rosendahl). In addition to these three clients, we formed two more clients, PAD-UFES \citep{pacheco2020pad} and Derm7pt \citep{kawahara2018seven}, following \cite{bdair2022semi}. These two clients are from the Federal University of Espirito Santo, Brazil (UFES$\_$brazil), and Simon Frazer University, Canada (SF$\_$canada). So, our setting contains a dataset of three countries in the northern hemisphere and two countries in the southern hemisphere. Due to the ozone hole, countries in the southern hemisphere may have more skin cancer cases than in the northern hemisphere \citep{li2022federated}. Then, the label distributions vary between clients. Therefore, our setting has a non-i.i.d problem that is likely in the real world.
We split the data of each client into training, validation, and test sets in the ratio of $7:1.5:1.5$. We used cross-entropy loss for the skin cancer classification task. Then, We also tested LN, but it showed consistently worse performance than BN or GN. We used the Adam optimizer, a batch size of $128$, a single local epoch, $300$ communication rounds, and $300$ total training epochs. In our experiments, we resized the input images to $256 \times 256 \times 3$. In the dataset, We tested the combinations of the number of local epochs and communication rounds, such as 5 \& 150 and 10 \& 30, but 1 \& 300 generally gave the best performance for all FL methods in all tasks. 

\paragraph{ECG dataset} 
\label{appendix:ecg_info}
As mentioned earlier, We used the PhysioNet 2021 \citep{reyna2021will} dataset and split it into five clients. The clients are Shaoxing People's Hospital, China (Shaoxing), CPSC 2018, China (CPSC), Georgia 12-lead Challenge Database, USA (Ga), Ningbo First Hospital, China (Ningbo), and Physikalisch-Technische Bundesanstalt, Germany (PTBXL).
Following \cite{oh2022lead}, we extracted samples with a sampling frequency $500$Hz and divided the data into $5$-second segments. Our aim is to solve a multi-label prediction task in which a $12$-lead ECG sample is given as input, and the objective is to diagnose $26$ types of cardiac diseases. We randomly split the data in the ratio of $8$:$1$:$1$ to form training, validation and test sets. Then, We used ResNet-NC-SE \citep{kang2022study} with asymmetric loss \citep{Ridnik_2021_ICCV} because it showed the best performance on 12-lead ECG readings in PhysioNet 2021 to the best of our knowledge \citep{torch_ecg_paper, torch_ecg}. We used the AdamW optimizer \citep{loshchilov2018decoupled}, a batch size of $64$, a single local epoch, $200$ communication rounds, and $200$ total training epochs. For the FedOpt-based methods, we only report the results of FedAdam for the ECG experiments because the other FedOpt-based methods, FedAdagrad and FedYoGi, showed similar performance, as in the eICU and skin cancer experiments. 

\section{Hyperparameters of FL methods} 
\label{appendix:hyperparameter}
In the section, We show the hyperparameters of FL methods. First of all, We use the same batch size for all FL methods in each dataset (256 in eICU, 128 in Skin cancer images, 64 in ECG). Then, the search space for other parameters is as follows :
\begin{itemize} 
\small
\item Learning rate($\eta$) := [0.1, 0.03, 0.01, 0.003, 0.001, 0.0001]
\item Mu($\mu$) := [1.0, 0.1, 0.01, 0.001, 0.0001]
\item Feddyn alpha($\alpha$) := [0.0001, 0.001, 0.01, 0.1]
\item Server learning rate($\eta_{g}$) := [0.1, 0.03, 0.01, 0.003, 0.001, 0.0001]
\item Tau($\gamma$) := [0.0001, 0.001, 0.01, 0.1] 
\end{itemize}
Also, the parameters to be tuned for each FL method are as follows :
\begin{itemize} 
\small
\item FedAvg, FedBN -- $\eta$
\item FedProx, FedPxN -- $\eta$, $\mu$
\item FedOpt -- $\eta$, $\eta_{g}$, $\gamma$
\item FedDyn -- $\eta$, $\alpha$
\end{itemize}

\section{More experimental results}
\label{appendix:extra_results}
In the section, we include more information and experimental results. (Table \ref{tab:eicu_5_clients_apurc} - \ref{tab:eicu_total_average_auprc})

\begin{table*}[h!]
\centering
\caption{AUPRC results for the eICU dataset using the data of the five largest clients. For each FL method, bold indicates the better normalization technique (LN or GN). We indicate the highest average AUPRC results for all six tasks in blue. }
\label{tab:eicu_5_clients_apurc}
\resizebox{\textwidth}{!}{
\begin{tabular}{c|ccccccccc} 
\toprule
\multicolumn{1}{l}{}              & \multicolumn{1}{l}{} & FedAvg                  & FedProx                 & FedBN                   & FedAdam                 & FedAdagrad              & FedYoGi                 & FedDyn                  & FedPxN                            \\ 
\hline
\multirow{2}{*}{mort\_24h}        & LN                   & \textbf{10.75$\pm$4.79} & \textbf{12.47$\pm$2.79} & 10.72$\pm$1.77          & \textbf{10.16$\pm$2.39} & \textbf{13.30$\pm$3.63} & 8.94$\pm$3.91           & \textbf{9.81$\pm$1.07}  & 13.07$\pm$2.77                    \\
                                  & GN                   & 10.58$\pm$1.03          & 12.29$\pm$2.15          & \textbf{11.42$\pm$3.23} & 10.06$\pm$1.75          & 12.46$\pm$1.32          & \textbf{10.39$\pm$2.70} & 8.27$\pm$2.39           & \textbf{15.64$\pm$0.19}           \\ 
\hline
\multirow{2}{*}{mort\_48h}        & LN                   & 12.92$\pm$0.49          & 13.14$\pm$1.77          & 12.66$\pm$1.18          & \textbf{14.50$\pm$0.95} & 12.78$\pm$0.80          & \textbf{13.35$\pm$0.55} & 10.13$\pm$2.75          & 13.13$\pm$1.29                    \\
                                  & GN                   & \textbf{13.84$\pm$0.46} & \textbf{13.20$\pm$2.09} & \textbf{14.07$\pm$0.44} & 14.15$\pm$0.94          & \textbf{15.77$\pm$1.38} & 12.90$\pm$2.54          & \textbf{14.81$\pm$2.84} & \textbf{13.90$\pm$1.41}           \\ 
\hline
\multirow{2}{*}{LOS}              & LN                   & 47.69$\pm$0.39          & 47.49$\pm$0.04          & 46.87$\pm$0.09          & 46.61$\pm$0.39          & 47.03$\pm$0.14          & 46.85$\pm$0.62          & 46.36$\pm$0.29          & \textbf{48.04$\pm$0.65}           \\
                                  & GN                   & \textbf{47.73$\pm$0.30} & \textbf{47.96$\pm$0.31} & \textbf{47.68$\pm$0.55} & \textbf{46.93$\pm$0.69} & \textbf{47.20$\pm$0.39} & \textbf{46.98$\pm$0.91} & \textbf{47.70$\pm$1.25} & 48.00$\pm$0.39                    \\ 
\hline
\multirow{2}{*}{disch\_24h}       & LN                   & \textbf{21.49$\pm$0.48} & \textbf{21.38$\pm$0.51} & \textbf{22.15$\pm$0.39} & 20.89$\pm$0.41          & 21.01$\pm$0.13          & 20.93$\pm$0.38          & 20.78$\pm$0.68          & 21.88$\pm$0.51                    \\
                                  & GN                   & ~21.39$\pm$0.56         & 21.19$\pm$0.20          & 21.59$\pm$0.77          & \textbf{21.00$\pm$0.32} & \textbf{21.12$\pm$0.53} & \textbf{21.09$\pm$0.53} & \textbf{22.06$\pm$0.19} & \textbf{22.21$\pm$0.22}           \\ 
\hline
\multirow{2}{*}{disch\_48h}       & LN                   & 22.03$\pm$1.04          & \textbf{21.35$\pm$0.13} & \textbf{22.56$\pm$1.18} & \textbf{22.99$\pm$1.52} & \textbf{23.02$\pm$2.21} & 22.58$\pm$1.48          & 21.13$\pm$0.44          & \textbf{22.44$\pm$2.06}           \\
                                  & GN                   & \textbf{22.17$\pm$1.01} & 21.17$\pm$0.02          & 21.76$\pm$1.21          & 22.92$\pm$1.59          & 22.85$\pm$1.12          & \textbf{23.10$\pm$1.61} & \textbf{21.65$\pm$0.85} & 22.40$\pm$1.57                    \\ 
\hline
\multirow{2}{*}{Acuity}           & LN                   & \textbf{21.11$\pm$0.30} & 21.28$\pm$0.06          & 21.22$\pm$0.20          & \textbf{20.87$\pm$0.79} & \textbf{20.70$\pm$0.79} & \textbf{20.85$\pm$1.35} & 21.03$\pm$0.21          & \textbf{21.46$\pm$0.45}           \\
                                  & GN                   & 20.97$\pm$0.13          & \textbf{21.54$\pm$0.44} & \textbf{21.31$\pm$0.16} & 20.59$\pm$1.20          & 19.46$\pm$0.30          & 20.57$\pm$1.32          & \textbf{21.94$\pm$0.48} & 21.34$\pm$0.40                    \\ 
\hline
\multirow{2}{*}{\textbf{Average}} & LN                   & 22.66                   & 22.85                   & 22.70                   & \textbf{22.67}          & 22.97                   & 22.25                   & 21.54                   & 23.33                             \\
                                  & GN                   & \textbf{22.78}          & \textbf{22.89}          & \textbf{22.97}          & 22.61                   & \textbf{23.14}          & \textbf{22.51}          & \textbf{22.74}          & \textbf{\textcolor{blue}{23.91}}  \\
\bottomrule
\end{tabular}
}
\end{table*}

\begin{table*}[h!]
\vspace{-3mm}
\centering
\caption{Average AUROC results for all six tasks for the eICU dataset using the data of the $5, 10, 20, 30$ largest clients. For each FL method and each client setting, bold indicates the better normalization (LN or GN). We indicate the highest AUROC for each client setting in blue. }
\vspace{-2mm}
\label{tab:eicu_total_average_auroc}
\begin{tabular}{c|cc|cc|cc|cc} 
\toprule
           & \multicolumn{2}{c|}{5 clients}                    & \multicolumn{2}{c|}{10 clients}                   & \multicolumn{2}{c|}{20 clients}          & \multicolumn{2}{c}{30 clients}                    \\ 
\cline{2-9}
           & LN             & GN                               & LN                               & GN             & LN    & GN                               & LN                               & GN              \\ 
\hline
FedAvg     & 67.88          & \textbf{68.01}                   & 67.34                            & \textbf{67.7}  & 68.62 & \textbf{68.87}                   & 67.52                            & \textbf{68.06}  \\
FedProx    & 69.44          & \textbf{69.85}                   & 68.66                            & \textbf{69.11} & 69.31 & \textbf{69.73}                   & 68.62                            & \textbf{68.94}  \\
FedBN    & 68.38          & \textbf{68.51}                   & \textbf{68.24}                   & 68.19          & 69.21 & \textbf{69.3}                    & \textbf{68.43}                   & 68.13           \\
FedAdam    & \textbf{67.67} & 67.42                            & 67.8                             & \textbf{67.87} & 68.09 & \textbf{68.96}                   & 67.67                            & \textbf{68.1}   \\
FedAdagrad & \textbf{67.38} & 67.31                            & 67.31                            & \textbf{67.68} & 67.98 & \textbf{68.85}                   & 67.3                             & \textbf{67.61}  \\
FedYoGi    & 65.97          & \textbf{67.96}                   & \textbf{67.67}                   & 66.95          & 68.38 & \textbf{68.6}                    & 67.66                            & \textbf{67.76}  \\
FedDyn     & 68             & \textbf{69.04}                   & 67.43                            & \textbf{67.46} & 67.29 & \textbf{67.89}                   & \textbf{67.26}                   & 66.14           \\
\mname    & 69.65          & \textbf{\textcolor{blue}{70.06}} & \textbf{\textcolor{blue}{69.55}} & 69.5           & 70.04 & \textbf{\textcolor{blue}{70.08}} & \textbf{\textcolor{blue}{69.22}} & 69.12           \\
\bottomrule
\end{tabular}

\vspace{-2mm}
\end{table*}

\begin{table*}[h!]
\centering
\caption{Average AUPRC results for all six tasks for the eICU dataset using the data of the $5, 10, 20, 30$ largest clients. For each FL method and each client setting, bold indicates the better normalization (LN or GN). We indicate the highest AUPRC for each client setting in blue. }
\label{tab:eicu_total_average_auprc}

\begin{tabular}{c|cc|cc|cc|cc} 
\toprule
           & \multicolumn{2}{c|}{5 clients}                    & \multicolumn{2}{c|}{10 clients}                   & \multicolumn{2}{c|}{20 clients}          & \multicolumn{2}{c}{30 clients}                     \\ 
\cline{2-9}
           & LN             & GN                               & LN             & GN                               & LN    & GN                               & LN             & GN                                \\ 
\hline
FedAvg     & 22.66          & \textbf{22.78}                   & 22.24          & \textbf{22.40}                   & 24.15 & \textbf{24.41}                   & 25.02          & \textbf{25.14}                    \\
FedProx    & 22.85          & \textbf{22.89}                   & 22.69          & \textbf{23.00}                   & 25.10 & \textbf{25.36}                   & 25.88          & \textbf{\textcolor{blue}{26.33}}  \\
FedBN      & 22.70          & \textbf{22.97}                   & \textbf{22.53} & 22.52                            & 24.16 & \textbf{24.87}                   & \textbf{25.26} & 25.06                             \\
FedAdam    & \textbf{22.67} & 22.61                            & 22.88          & \textbf{22.99}                   & 24.55 & \textbf{25.28}                   & 24.75          & \textbf{25.72}                    \\
FedAdagrad & 22.97          & \textbf{23.14}                   & 22.42          & \textbf{22.91}                   & 23.89 & \textbf{25.22}                   & 25.03          & \textbf{25.69}                    \\
FedYoGi    & 22.25          & \textbf{22.51}                   & \textbf{22.80} & 22.01                            & 24.31 & \textbf{24.48}                   & \textbf{25.26} & 25.05                             \\
FedDyn     & ~21.54         & \textbf{22.74}                   & \textbf{22.47} & 21.73                            & 23.81 & \textbf{23.93}                   & \textbf{24.73} & 24.48                             \\
FedPxN     & 23.33          & \textbf{\textcolor{blue}{23.91}} & 23.28          & \textbf{\textcolor{blue}{23.39}} & 25.30 & \textbf{\textcolor{blue}{25.72}} & 26.06          & \textbf{26.29}                    \\
\bottomrule
\end{tabular}
\end{table*}

\vspace{-3mm}












\end{document}